\documentclass[acmsmall,screen]{acmart}

\AtBeginDocument{%
  \providecommand\BibTeX{{%
    \normalfont B\kern-0.5em{\scshape i\kern-0.25em b}\kern-0.8em\TeX}}}

\setcopyright{acmcopyright}
\copyrightyear{2018}
\acmYear{2018}
\acmDOI{10.1145/1122445.1122456}

\usepackage{booktabs}  
\usepackage{MnSymbol}  
\usepackage{multirow}  
\usepackage{algorithm}  
\usepackage{algpseudocode}

\acmJournal{JACM}
\acmVolume{37}
\acmNumber{4}
\acmArticle{111}
\acmMonth{8}

\begin{document}

\title{Auto IV: Counterfactual Prediction via Automatic Instrumental Variable Decomposition}

\author{Junkun Yuan$^\dagger$}
\orcid{0000-0003-0012-7397}
\email{yuanjk@zju.edu.cn}
\affiliation{%
  \institution{Zhejiang University}
  \city{Zhejiang}
  \country{China}
}

\author{Anpeng Wu$^\dagger$}
\email{anpwu@zju.edu.cn}
\affiliation{%
  \institution{Zhejiang University}
  \city{Zhejiang}
  \country{China}
}

\author{Kun Kuang}
\authornote{Corresponding author.}
\email{kunkuang@zju.edu.cn}
\affiliation{%
  \institution{Zhejiang University}
  \city{Zhejiang}
  \country{China}
}

\author{Bo Li}
\email{libo@sem.tsinghua.edu.cn}
\affiliation{%
  \institution{Tsinghua University}
  \city{Beijing}
  \country{China}
}

\author{Runze Wu}
\email{wurunze1@corp.netease.com}
\affiliation{%
  \institution{NetEase Fuxi AI Lab}
  \city{Zhejiang}
  \country{China}
}

\author{Fei Wu}
\email{wufei@cs.zju.edu.cn}
\affiliation{%
  \institution{Zhejiang University}
  \city{Zhejiang}
  \country{China}
}

\author{Lanfen Lin}
\email{llf@zju.edu.cn}
\affiliation{%
  \institution{Zhejiang University}
  \city{Zhejiang}
  \country{China}
}

\renewcommand{\shortauthors}{Trovato and Tobin, et al.}

\begin{abstract}
Instrumental variables (IVs), sources of treatment randomization that are conditionally independent of the outcome, play an important role in causal inference with unobserved confounders. However, the existing IV-based counterfactual prediction methods need well-predefined IVs, while it’s an art rather than science to find valid IVs in many real-world scenes. Moreover, the predefined hand-made IVs could be weak or erroneous by violating the conditions of valid IVs. These thorny facts hinder the application of the IV-based counterfactual prediction methods. In this paper, we propose a novel Automatic Instrumental Variable decomposition (AutoIV) algorithm to automatically generate representations serving the role of IVs from observed variables (IV candidates). Specifically, we let the learned IV representations satisfy the relevance condition with the treatment and exclusion condition with the outcome via mutual information maximization and minimization constraints, respectively. We also learn confounder representations by encouraging them to be relevant to both the treatment and the outcome. The IV and confounder representations compete for the information with their constraints in an adversarial game, which allows us to get valid IV representations for IV-based counterfactual prediction. Extensive experiments demonstrate that our method generates valid IV representations for accurate IV-based counterfactual prediction.
\end{abstract}

\ccsdesc[500]{Computing methodologies~Causal reasoning and diagnostics}
\ccsdesc[300]{Computing methodologies~Machine learning}
\ccsdesc[300]{Computing methodologies~Statistical relational learning}

\keywords{instrumental variable, counterfactual prediction, causal inference, representation learning, mutual information.}

\maketitle

\section{Introduction}\label{sec:introduction}
As a representative task in machine learning \cite{roh2019survey, frunza2010machine, gao2018declarative, karpatne2018machine}, supervised learning \cite{dam2007neural, wang2016towards} explores correlations between variables from rich data for prediction. However, in many real applications, a decision-maker always wants to judge the counterfactual impact of treatment (policy) changes on the outcome that can not be found in the data. For example, an airline wants to estimate the effect of prices (i.e. treatment) on customers' purchase tendency (i.e. outcome) \cite{jason2017deepiv}. We may observe that examples with high prices are often associated with high sales in data sampled during holidays, which may fool the direct supervised learning approaches to predict that increasing prices would also lead to high sales at other times. In this case, we can add the observable confounders (i.e., holidays, which cause the changes in both the prices and the sales) into training data to correct the model. Nevertheless, if there exist unobserved confounders (e.g., conferences, which are also common causes of the prices and the sales but are unknown to the decision-maker), the typical supervised learning model would still head in the wrong direction. 

Instrumental Variables (IVs) \cite{wright1928tariff} are exogenous variables that are correlated to the treatment but do not directly affect the outcome, which provides an alternative approach for counterfactual prediction even with the unobserved confounders. Existing IV-based counterfactual prediction methods mainly adopt a two-stage procedure, which first builds a model to estimate the treatment based on the IVs, and then predicts the outcome with the estimated treatment. Two-stage least squares (2SLS) \cite{angrist2008mostly} is a well-known method that employs the two-stage procedure with linear models and obtains homogeneous treatment effects. Recent IV-based counterfactual prediction works \cite{jason2017deepiv, lewis2018adversarial, muandet2019dual, singh2019kernel, bennett2019deep} mainly focus on generalizing previous approaches on high-dimensional and non-linear data. These methods achieve great counterfactual prediction performance, however, they rely heavily on well-predefined IVs. In many real-world applications, we can hardly have enough prior knowledge to identify the valid IVs \cite{kuang2020ivy} (i.e., the variables that satisfy the relevance, the exclusion, and the unconfounded instrument conditions, see Sec. \ref{sec-pre} for details). Moreover, the predefined hand-made IVs could be weak or erroneous by violating some of the conditions of the valid IVs. Therefore, it's highly demanding to develop a data-driven approach to automatically obtain valid IVs (or IV representations) for the downstream IV-based counterfactual prediction methods. 

In many real applications, although there are always a large number of observed variables, few of them satisfy the conditions of the valid IVs. Since finding the valid IVs is difficult, instead, there are growing works that focus on synthesizing valid summary IVs with IV candidates \cite{kuang2020ivy} (some of them might be invalid IVs, i.e., do not strictly satisfy the conditions of the valid IVs). Mendelian Randomization (MR) \cite{bowden2015mendelian} is a popular approach that utilizes genetic markers as the IVs to perform causal inference \cite{Yao2021ASO} among clinical factors. Unweighted/Weighted Allele Scores (UAS/WAS) \cite{burgess2013use, davies2015many, burgess2016combining} that weigh each IV candidate equally or based on the correlation between them and the treatment are representative methods in MR. However, they need all the IV candidates to be both valid and independent conditional on the summary IVs. Hartford et al. \cite{hartford2020valid} apply an ensemble method to select a valid IV set with asymptotical validity. But it not only relies on the independence and modal validity of IV candidates but also needs high computation costs by running the downstream IV-based methods with every IV candidate for valid set selection. Kuang et al. \cite{kuang2020ivy} present to model a summary IV as a latent variable and estimate it by utilizing recent advances in weak supervision that is based on statistical dependencies among the IV candidates. However, this method is confined to the binary variable setting, limiting its use in many real-world applications. 

Inspired by the recent works \cite{yao2019estimation, hassanpour2020learning, wu2020learning} on causal disentangled representation learning, we argue that although invalid IV candidates do not satisfy the conditions of the valid IVs strictly, one might decompose and utilize a part of their information to generate IV representations.
Therefore, in this paper, we propose a novel Automatic Instrumental Variable decomposition (AutoIV) algorithm to automatically generate representations serving the role of IVs for counterfactual prediction with fewer constraints for the IV candidates. Specifically, we first generate the IV representations from the IV candidates and make them satisfy the relevance condition with the treatment and the exclusion condition with the outcome via mutual information maximization and minimization constraints, respectively. We also generate confounder representations by encouraging them to be relevant to both the treatment and the outcome. The IV and the confounder representations compete for the corresponding information with their constraints in an adversarial game, which allows us to obtain valid IV representations for counterfactual prediction with the downstream IV-based methods. 

In summary, the main contributions of this paper are: 
\begin{itemize}
\item We study the problem of IV-based counterfactual prediction under a more practical setting, i.e., no valid IVs are available for learning, which is beyond the capability of the previous IV-based methods.
\item We propose a novel Automatic Instrumental Variable decomposition (AutoIV) algorithm to automatically generate IV representations that satisfy the conditions of the valid IVs from the IV candidates. It adopts mutual information constraints to control representation learning process via an adversarial game.
\item Extensive experiments show that the proposed method generates valid IV representations for accurate counterfactual prediction, which is even comparable to directly using the true valid IVs.
\end{itemize}

The rest of the paper is organized as follows. 
In Sec. \ref{sec-rel}, some related works about IV-based counterfactual prediction, IV synthesis, and causal representation learning are introduced. 
In Sec. \ref{sec-pre}, the definition of the valid IVs and some related IV-based methods are stated. 
In Sec. \ref{sec-met}, our automatic instrumental variable decomposition algorithm is introduced. 
In Sec. \ref{sec-exp}, the results of the experiments on low-dimensional and high-dimensional are reported.
We discuss the investigation with a future research outlook in Sec. \ref{sec-con}.

\section{Related Work}\label{sec-rel}
In this section, we briefly review the related works of IV-based counterfactual prediction, IV synthesis, and causal representation learning in recent years.

\subsection{IV-based Counterfactual Prediction}
Two-stage least squares (2SLS) \cite{angrist2008mostly} is a representative method for IV-based counterfactual prediction with linear models in causal inference researches \cite{Yao2021ASO, Kuang2020TreatmentEE, Amornbunchornvej2021VariablelagGC, Yu2021AUV, kuang2020data}. Many recent IV-based counterfactual prediction methods extend 2SLS to non-linear and high-dimensional settings. One research direction is the generalized method of moments (GMM) \cite{hansen1982large}, which uses moment conditions to estimate model parameters. A recent trend is to combine GMM with machine learning, like selecting moment conditions via adversarial training \cite{lewis2018adversarial} and variational reformulation of GMM with deep neural networks \cite{bennett2019deep}. Another direction is based on kernel approaches, such as a single-stage kernel approach \cite{muandet2019dual} and a novel method with consistency guarantees \cite{singh2019kernel}. DeepIV \cite{jason2017deepiv} is a recent remarkable study that fits a mixture density network for the treatment and trains an outcome prediction model with the estimated conditional treatment distribution. All of the above methods need predefined IVs, and their performance relies on the validity of the given IVs. However, identifying and obtaining valid IVs may be thorny because their validity conditions are strict.

\subsection{IV Synthesis}
There are growing works \cite{bowden2016consistent, bowden2015mendelian, kang2016instrumental, bowden2016consistent, windmeijer2019use, han2008detecting, hartford2020valid, kuang2020ivy} that propose to synthesize a valid summary IV by using the given observed variables (IV candidates) in recent years. Among them, some works \cite{bowden2016consistent, bowden2015mendelian} are based on the independence condition of IV candidates, which is a strong restrictive property \cite{kuang2020ivy}. Some approaches \cite{kang2016instrumental, bowden2016consistent, windmeijer2019use, han2008detecting} perform reliable estimation only when most of the IV candidates are valid, which is also a strong condition. Hartford et al. \cite{hartford2020valid} adopt ensemble methods based on the modal validity of the IV candidates, however, it needs expensive computation cost to select the valid IV set. 
Unweighted/Weighted Allele Scores (UAS/WAS) \cite{burgess2013use, davies2015many, burgess2016combining} weigh each IV candidate equally or based on the correlation between them and the treatment. 
Kuang et al. \cite{kuang2020ivy} generalize the allele scores method \cite{burgess2013use,davies2015many,burgess2016combining}, which builds a summary IV and estimates it with advanced methods from weak supervision and structure learning. However, it only applies to the binary variable setting. 
These previous IV synthesis methods rely on some strong conditions for the IV candidates and may not be practical in many real scenes, while we present an automatic IV representation learning algorithm that only needs mild assumptions in this paper. Take the airline case as an example. When we are looking for valid IVs, e.g., fuel costs, from the IV candidates, we do not need to assume that they are valid, modal validity, or binary, but only need them to be correlated with the treatment, i.e., price, and be independent of the unobserved confounders, i.e., conferences.

\subsection{Causal Representation Learning}
Recently, causal representation learning \cite{johansson2016learning, shalit2017estimating, kuang2017estimating, yao2019estimation, hassanpour2020learning, wu2020learning, kuang2020data} has attracted lots of attention in many applications \cite{kuang2018stable, wang2021causal, Yang_2021_CVPR, yue2021transporting, niu2021counterfactual, kuang2021balance}. 
Among these works, Yao et al. \cite{yao2019estimation} propose to reduce prediction bias by filtering out the nearly IVs. 
Some works \cite{hassanpour2020learning, wu2020learning} decompose the IV, confounder, and adjustment representations by encouraging or limiting the correlations between variables. 
However, these works are limited to the binary treatment setting. Moreover, they neither give empirical results to show the effectiveness of the learned IV representations nor make use of the decomposed IV representations for counterfactual prediction. 
In contrast, we present a data-driven IV representation learning algorithm and show its effectiveness by applying the learned representations to the downstream IV-based methods for accurate counterfactual prediction.

\section{Preliminary}\label{sec-pre}
By following previous works \cite{bennett2019deep, singh2019kernel}, we assume the relationship between treatment variable $X$ and outcome variable $Y$ in data generating process is
\begin{equation}\label{equ-gen}
    Y = g(X) + e,
\end{equation} 
where $g(\cdot)$ is an unknown causal response function which is potentially non-linear and continuous, and $e$ is the error term that contains unobserved latent factors (i.e. unmeasured confounders) which affect both $X$ and $Y$. Here, we assume the error term $e$ is with zero expectation and finite variance (i.e., $\mathbb{E}[e]=0$ and $\mathbb{E}[e^2]<\infty$). $e$ contains unobserved factors that affect $X$, thus $e$ would be correlated with $X$, i.e. $\mathbb{E}[e|X]\neq 0$, which makes $X$ an endogenous variable and leads to $g(X)\neq\mathbb{E}[Y|X]$. Thus, it is infeasible to estimate the causal relationship $g(\cdot)$ between $X$ and $Y$ via directly estimating $\mathbb{E}[Y|X]$ from data distribution $P(X,Y)$ because of the confounding effect caused by the unobserved error $e$. 
The instrumental variables (IVs) are introduced to solve the endogenous treatment problem as we introduced previously. Valid IVs (denoted by $Z$) should satisfy the following conditions \cite{jason2017deepiv, singh2019kernel, bennett2019deep}: 
\begin{itemize}
    \item \textbf{Relevance.} $Z$ is related to $X$, i.e., $\mathbb{P}(X|Z)\neq\mathbb{P}(X)$;
    \item \textbf{Exclusion.} $Z$ does not directly affect $Y$, i.e., $\mathbb{P}(Y|Z,X,e)=\mathbb{P}(Y|X,e)$;
    \item \textbf{Unconfounded Instrument.} $Z$ should be unconfounded, i.e., $\mathbb{E}[e|Z]=\mathbb{E}[e]$.
\end{itemize}
The goal of IV-based counterfactual prediction is to obtain a counterfactual estimation function $\hat{g}$ that is close to the true response function $g$. 
Moreover, if there exists exogenous variable $C$ (i.e., $\mathbb{P}(e|C)=\mathbb{P}(e)$), we can make use of it for more accurate estimation, i.e. $X=(X',C)$ and $Z=(Z',C)$, where $X'$ and $Z'$ are the true treatment variable and instrumental variable, respectively. 
Note that we will also learn confounder representations in our algorithm, which are used as the exogenous variables $C$ in the IV-based counterfactual prediction process.

Previous IV-based counterfactual prediction approaches assume that they have access to the true valid IVs $Z$ which strictly satisfy the above conditions. Then, we could identify the causal response function $g(\cdot)$ based on
\begin{equation}
    \mathbb{E}[Y|Z]=\mathbb{E}[g(X)|Z]=\int{g(X)d\mathbb{P}(X|Z)}.
\end{equation}
That is, one may first learn $\mathbb{P}(X|Z)$, then use it to estimate $g(\cdot)$. For example, standard two-stage least squares (2SLS) method \cite{angrist2008mostly} first learns $\mathbb{E}[\phi(X)|Z]$ with linear basis $\phi(\cdot)$, then fits $Y$ by least-squares regression with the coefficient $\hat{\phi}(\cdot)$ that estimated in the first stage. Some non-parametric works \cite{newey2003instrumental, darolles2011nonparametric} extend the model basis to more complicated mapping functions or regularization, e.g. polynomial basis. DeepIV \cite{jason2017deepiv} is proposed to apply deep neural networks in the two-stage procedure. It fits a mixture density network $F_{\phi}(X|Z)$ in the first stage and regresses $Y$ by sampling from the estimated mixture Gaussian distributions of $X$. KernelIV \cite{singh2019kernel} is a recent kernel approach that maps $Z$, $X$, and $Y$ to reproducing kernel Hilbert spaces and perform the two-stage procedure in that space. DeepGMM \cite{bennett2019deep} extends the existing GMM methods in the high-dimensional treatment and IVs setting, which is based on a novel variational reformulation of the optimally-weighted GMM. 

The above existing IV-based counterfactual prediction methods need well-predefined valid IVs. However, it is an art rather than science to find suitable IVs in real applications. Even worse, the predefined hand-made IVs could be weak or erroneous by violating the conditions. Without the valid IVs, the counterfactual prediction performance of these downstream IV-based methods cannot be guaranteed.

In this paper, we aim to automatically learn valid IV representations that can be applied to the downstream IV-based methods for accurate counterfactual prediction. The validity of the learned IV representation determines the accuracy of the downstream counterfactual prediction task.

\section{Method}\label{sec-met}
\begin{figure}[t]
    \centering
    \includegraphics[trim={0cm 0cm 0cm 0cm},width=0.2\columnwidth]{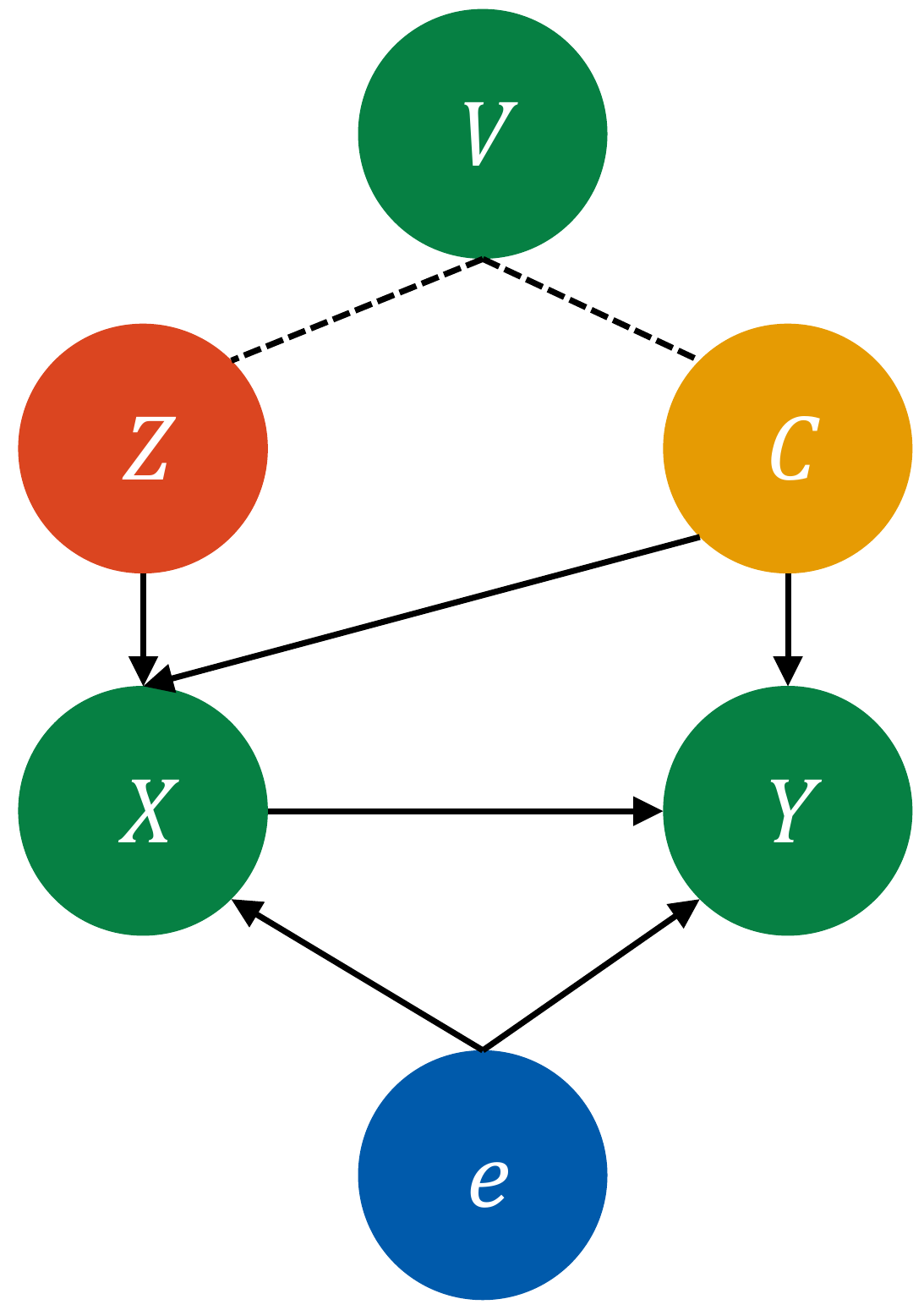}
    \caption{
    The proposed AutoIV framework. Variables $V$, $X$, and $Y$ are corresponding to the observed variables, treatment, and outcome, respectively. Variables $e$ are unobserved confounders that are related to both 
    $X$ and $Y$. AutoIV decomposes representations of instrumental variables $Z$ and confounders $C$ from the observed variables $V$ automatically, then use the learned representations for IV-based counterfactual prediction.}
    \label{fig-framework}
\end{figure}

In this section, we propose a novel Automatic Instrumental Variable decomposition (AutoIV) algorithm to generate decomposed IV and confounder representations from the observed variables. The proposed framework of AutoIV is shown in Fig. \ref{fig-framework}. The green part represents all the available variables, including observed variables $V$, treatment variables $X$, and outcome variables $Y$. $E$ denotes unobserved confounders that are related to both $X$ and $Y$. The observed variables $V$ are correlated with $X$ and also might be associated with $Y$. Similar to the general setting in recent IV analysis works \cite{jason2017deepiv, singh2019kernel, bennett2019deep}, observed variables $V$ are assumed exogenous, i.e. $\mathbb{P}[E|V]=\mathbb{P}[E]$. Therefore, the decomposed representations of instrumental variables $Z$ and confounders $C$ are also exogenous, which satisfies the unconfounded IV condition. Suppose that we have data $\mathcal{D}=\{(\boldsymbol{v}_{i},\boldsymbol{x}_{i},\boldsymbol{y}_{i})\}_{i=1}^N$, our goal is to learn the representations of $Z$ and $C$ from the observed variables $V$ based on their relationships to $X$ and $Y$ with data $\mathcal{D}$. Then, we use the learned representations for counterfactual prediction with the downstream IV-based methods introduced in Sec. \ref{sec-pre}. The validity of the learned representations determines the accuracy of the IV-based counterfactual prediction.

We first use neural networks to model the representations for $Z$ and $C$ as $\phi^Z(\cdot)$ and $\phi^C(\cdot)$ with parameters $\theta_{\phi^Z}$ and $\theta_{\phi^C}$, respectively. The observed variables $V$ are used as inputs of the representation networks. We control the information that flows into $\phi^Z(\cdot)$ to be related to $X$ and conditionally independent of $Y$, which is based on the relevance and exclusion conditions, respectively. We then let $\phi^C(\cdot)$ be related to both $X$ and $Y$. These two representation networks compete for the corresponding information with their constraints in an adversarial game. A general two-stage counterfactual prediction loss is then employed to further calibrate the learned representations. 

Let $A$ and $B$ be two random variables that are correlated with each other. We have examples $\boldsymbol{a}_{i}$ and $\boldsymbol{b}_{i}$ sampled from the distributions of $A$ and $B$, respectively. 
We encourage (or discourage) the relevance between $A$ and $B$ by maximizing (or minimizing) the mutual information between them. However, only the samples $\{(\boldsymbol{a}_{i},\boldsymbol{b}_{i})\}_{i=1}^N$ are available in our task, but what mutual information estimation needs is data distributions. 
Inspired by recent works on contrastive learning and sample-based mutual information estimation \cite{oord2018representation, cheng2020club}, we first learn a variational distribution $q(B|A)$ to approximate $\mathbb{P}(B|A)$. We let positive sample pair to be the sample pair with the same index (i.e. $(a_{i},b_{i})$), and let negative sample pair be the sample pair with the different index $(a_{i},b_{j})_{i\neq j}$. As we already have the variational approximation $q(B|A)$, we can increase (or decrease) the relevance between $A$ and $B$ by maximizing (or minimizing) the differences between the variational approximation of the positive sample pair (i.e. $q(b_{i}|a_{i})$) and that of the negative sample pair (i.e. $q(b_{j}|a_{i})$). It can intuitively be interpreted that mutual information maximization task is achieved when there exist distinct differences between the relevance of $a_{i}$ to its corresponding $b_{i}$ and the relevance of $a_{i}$ to $b_{j}$ (where $i\neq{j}$). Meanwhile, mutual information minimization is to reduce that differences. Although there is deviation between $q(B|A)$ and $\mathbb{P}(B|A)$, the estimated mutual information is still excellent with great variational approximation \cite{cheng2020club}.

\subsection{Learning IV Representations}
We aim to learn the IV representations that satisfy the conditions of the valid IVs (see Sec. \ref{sec-pre}), i.e., relevance, exclusion, and unconfounded instrument. Since we have already assumed the exogeneity of the observed variables $V$ by following previous works \cite{jason2017deepiv, singh2019kernel, bennett2019deep}, and the learned representations always satisfy the unconfounded instrument condition, we only need to make the learned IV representations satisfy the relevance condition with the treatment and the exclusion condition with the outcome.  

\textbf{Learning relevance.} 
The relevance condition, i.e., $\mathbb{P}(X|Z)\neq\mathbb{P}(X)$, requires IV representations $\phi^Z(V)$ to be correlated with the treatment $X$. Therefore, we encourage the information of $V$ that is related to $X$ to enter the IV representations $\phi^Z(V)$. 
We first use variational distribution $q_{\theta_{ZX}}(X|\phi^Z(V))$ with neural network parameters $\theta_{ZX}$ to approximate the true conditional distribution $\mathbb{P}(X|\phi^Z(V))$. The log-likelihood loss function of variational approximation $q_{\theta_{ZX}}(X|\phi^Z(V))$ with $N$ samples is given as:
\begin{equation}\label{equ-zx-lld}
    \mathcal{L}_{ZX}^{LLD}=-\frac{1}{N}\sum_{i=1}^N{\log{q_{\theta_{ZX}}(\boldsymbol{x}_i|\phi^Z(\boldsymbol{v}_i))}}.
\end{equation}
We minimize Eq. (\ref{equ-zx-lld}) to get optimal variational approximation $q_{\hat{\theta}_{ZX}}(X|\phi^Z(V))$ with parameters $\hat{\theta}_{ZX}$. 
To increase the relevance between the IV representations and the treatment, we maximize the mutual information between them with
\begin{equation}\label{equ-zx-mi}
    \begin{aligned}
        \mathcal{L}_{ZX}^{MI}=-\frac{1}{N^2}\sum_{i=1}^N\sum_{j=1}^N(\log{q_{\theta_{ZX}}(\boldsymbol{x}_{i}|\phi^Z(\boldsymbol{v}_i))}-
        \log{q_{\theta_{ZX}}(\boldsymbol{x}_{j}|\phi^Z(\boldsymbol{v}_i))}),
    \end{aligned}
\end{equation}
where $\log{q_{\theta_{ZX}}(\boldsymbol{x}_{i}|\phi^Z(\boldsymbol{v}_i)})$ represents the conditional log-likelihood of positive sample pair $(\phi^Z(\boldsymbol{v}_{i}),\boldsymbol{x}_{i})$ and $q_{\theta_{ZX}}(\boldsymbol{x}_{j}|\phi^Z(\boldsymbol{v}_i)_{i\neq j}$ represents the negative sample pair $(\phi^Z(\boldsymbol{v}_{i}),\boldsymbol{x}_{j})_{i\neq j}$. We minimize Eq.(\ref{equ-zx-mi}) to optimize the IV representations $\phi^Z(V)$ for relevance condition via maximizing differences between the positive and negative sample pairs. 

\textbf{Learning exclusion.} The exclusion condition requires IV representations to be related to the outcome $Y$ only through the treatment $X$ and unobserved error $e$, i.e. $\mathbb{P}(Y|Z,X,e)=\mathbb{P}(Y|X,e)$. 
Since $e$ is unobserved, we employ a more strict condition instead, i.e., $Z\upmodels Y|X$. 
Therefore, we minimize mutual information between $Z$ and $Y$ conditional on $X$. Similarly, we first use variational distribution $q_{\theta_{ZY}}(Y|\phi^Z(V))$ with parameters $\theta_{ZY}$ to approximate the true conditional distribution $\mathbb{P}(Y|\phi^Z(V))$. The log-likelihood loss function for $q_{\theta_{ZY}}(Y|\phi^Z(V))$ is given as
\begin{equation}\label{equ-zy-lld}
    \mathcal{L}_{ZY}^{LLD}=-\frac{1}{N}\sum_{i=1}^N{\log{q_{\theta_{ZY}}(\boldsymbol{y}_i|\phi^Z(\boldsymbol{v}_i))}}.
\end{equation}
The optimal variational approximation $q_{\hat{\theta}_{ZY}}(\boldsymbol{y_i}|\phi^Z(\boldsymbol{v_i}))$ is achieved with parameters $\hat{\theta}_{ZY}$ by minimizing Eq. (\ref{equ-zy-lld}). 
The IV representations $\phi^{Z}(V)$ should be independent of the outcome $Y$ given the treatment $X$, we achieve it by minimizing the mutual information between them. Since the treatments $X$ are continuous random variables, we consider the constraints of conditional independence with smooth weight $w_{ij}$, and the loss function for mutual information minimization between IV representations $\phi^{Z}(V)$ and the outcome $Y$ is given as:
\begin{equation}\label{equ-zy-mi}
    \begin{aligned}
        \mathcal{L}_{ZY}^{MI}=\frac{1}{N^2}\sum_{i=1}^N\sum_{j=1}^N(\omega_{ij}\cdot(\log{q_{\theta_{ZY}}(\boldsymbol{y}_{i}|\phi^Z(\boldsymbol{v}_i))}-
        \log{q_{\theta_{ZY}}(\boldsymbol{y}_{j}|\phi^Z(\boldsymbol{v}_i))})).
    \end{aligned}
\end{equation}
Different from mutual information maximization in learning relevance, we let the positive ($(\phi^Z(\boldsymbol{v}_{i}),\boldsymbol{y}_{i})$) and negative $(\phi^Z(\boldsymbol{v}_{i}),\boldsymbol{y}_{j})$ sample pairs have close a log-likelihood expectation to make the IV representations $\phi^{Z}(V)$ and the outcome $Y$ conditional independent. $\omega_{ij}$ is the weight of each pair of positive and negative samples, and we determine it by the discrepancy between $\boldsymbol{x}_{i}$ and $\boldsymbol{x}_{j}$ in RBF kernel:
\begin{equation}
    \omega_{ij}={\rm{softmax}}(e^{-\frac{{\Vert \boldsymbol{x}_{i}-\boldsymbol{x}_{j} \Vert}^2}{2\sigma ^2}}), \quad i,j={1,2,...,N},
\end{equation}
where $\sigma$ is a hyperparameter, we use 0.5 for it in our experiments. The weight of positive and negative sample pairs increases when their treatments $X$ have closer distance. In other words, we would like to pay attention to the pairs which have close $X$ values for our conditional independent constraints.

\subsection{Learning Confounder Representations}
We also decompose and learn the representations of confounders that are correlated to both the treatment and outcome. They are used as exogenous variables $C$ for counterfactual prediction (see Sec. \ref{sec-pre}). We let the generated confounder representations, i.e. $\phi^C(V)$, are both correlated to the treatment $X$ and outcome $Y$ variables. With the similar procedure in learning IV representations, we first use variational distribution $q_{\theta_{CX}}(X|\phi^C(V))$ to approximate conditional distribution $\mathbb{P}(X|\phi^C(V))$, and the corresponding log-likelihood loss function is given as:
\begin{equation}\label{equ-cx-lld}
    \mathcal{L}_{CX}^{LLD}=-\frac{1}{N}\sum_{i=1}^N{\log{q_{\theta_{CX}}(\boldsymbol{x}_i|\phi^C(\boldsymbol{v}_i))}},
\end{equation}
Optimal approximation $q_{\hat{\theta}_{CX}}(X|\phi^C(V))$ with parameter $\hat{\theta}_{CX}$  is obtained by minimizing (\ref{equ-cx-lld}). We then minimize the loss function of mutual information maximization between confounder representations $\phi^{C}(V)$ and the treatment $X$:
\begin{equation}\label{equ-cx-mi}
    \begin{aligned}
        \mathcal{L}_{CX}^{MI}=-\frac{1}{N^2}\sum_{i=1}^N\sum_{j=1}^N(\log{q_{\theta_{CX}}(\boldsymbol{x}_{i}|\phi^C(\boldsymbol{v}_i))}-
        \log{q_{\theta_{CX}}(\boldsymbol{x}_{j}|\phi^C(\boldsymbol{v}_i))}).
    \end{aligned}
\end{equation}
The pairs of positive sample $(\phi^C(\boldsymbol{v}_i),\boldsymbol{x}_i)$ and negative sample $(\phi^C(\boldsymbol{v}_i),\boldsymbol{x}_j)$ are used to increase the relevance between $C$ and $X$. Also, the variational distribution $q_{\theta_{CY}}(Y|\phi^C(V)$ for conditional distribution $\mathbb{P}(Y|\phi^C(V))$ and its mutual information maximization loss function is given as: 
\begin{equation}\label{equ-cy-lld}
    \mathcal{L}_{CY}^{LLD}=-\frac{1}{N}\sum_{i=1}^N{\log{q_{\theta_{CY}}(\boldsymbol{y}_i|\phi^C(\boldsymbol{v}_i))}},
\end{equation}
\begin{equation}\label{equ-cy-mi}
    \begin{aligned}
        \mathcal{L}_{CY}^{MI}=-\frac{1}{N^2}\sum_{i=1}^N\sum_{j=1}^N(\log{q_{\theta_{CY}}(\boldsymbol{y}_{i}|\phi^C(\boldsymbol{v}_i))}-
        \log{q_{\theta_{CY}}(\boldsymbol{y}_{j}|\phi^C(\boldsymbol{v}_i))}).
    \end{aligned}
\end{equation}
We minimize Eq. (\ref{equ-cy-lld}) to get optimal variational approximation $q_{\hat{\theta}_{CY}}(Y|\phi^C(V))$ with parameter $\hat{\theta}_{CY}$, and minimize Eq. (\ref{equ-cy-mi}) to encourage the confounder representations $\phi^{C}(V)$ and the outcome $Y$ to be relevant. 

Since conditional on the confounders that contain IV information would introduce bias in causal inference \cite{wooldridge2016should}, also, if the information of confounders (i.e. the variables correlated to $Y$) is embedded in the IV representations would influence the exclusion condition. 
Therefore, we minimize mutual information between the IV representations $\phi^Z(V)$ and confounder representations $\phi^C(V)$
to regularize the learned information in the generated representations. 
The variational distribution $q_{\theta_{ZX}}(\phi^C(V)|\phi^Z(V)$ for conditional distribution $\mathbb{P}(\phi^C(V)|\phi^Z(V))$ and the mutual information minimization loss function are given as:
\begin{equation}\label{equ-zc-lld}
    \mathcal{L}_{ZC}^{LLD}=-\frac{1}{N}\sum_{i=1}^N{\log{q_{\theta_{ZC}}(\phi^C(\boldsymbol{v}_i)|\phi^Z(\boldsymbol{v}_i))}},
\end{equation}
\begin{equation}\label{equ-zc-mi}
    \begin{aligned}
        \mathcal{L}_{ZC}^{MI}=\frac{1}{N^2}\sum_{i=1}^N\sum_{j=1}^N(\log{q_{\theta_{ZC}}(\phi^C(\boldsymbol{v}_i)|\phi^Z(\boldsymbol{v}_i))}-
        \log{q_{\theta_{CY}}(\phi^C(\boldsymbol{v}_j)|\phi^Z(\boldsymbol{v}_i))}).
    \end{aligned}
\end{equation}
We minimize Eq. (\ref{equ-zc-lld}) to learn accurate variational approximation $q_{\theta_{ZX}}(\phi^C(V)|\phi^Z(V)$ for the conditional distribution $\mathbb{P}(\phi^C(V)|\phi^Z(V))$, and use the variational approximation to regularize the IV and confounder representations via minimizing Eq. (\ref{equ-zc-mi}).

In the above procedure with mutual information constraints, the IV representations $\phi^Z(V)$ attempt to extract information that is correlated to the treatment $X$ and conditional independent to the outcome $Y$, while the confounder representations $\phi^C(V)$ are encouraged to be correlated to both $X$ and $Y$. We also employ a regularization term to encourage the information to enter one of the extracted representations. Therefore, the two representation networks compete for the corresponding information with their constraints in an adversarial game, which allows us to get valid IV and confounder representations. We then introduce the general IV-based counterfactual prediction procedure to further improve the learned representations in the following.

\subsection{Representation Calibration}
We combine mutual information-based representation learning with a general two-stage counterfactual prediction procedure to further calibrate the learned representations. More concretely, we first regress $X$ on IV and confounder representations, i.e., $\phi^Z(V)$ and $\phi^C(V)$,
\begin{equation}\label{equ-x}
    \mathcal{L}_{X}=\frac{1}{N}\sum_{i=1}^Nl(\boldsymbol{x}_i,f^X(\phi^Z(\boldsymbol{v}_i),\phi^C(\boldsymbol{v}_i))),
\end{equation}
where $f^X$ is the first-stage (treatment) regression network with parameter $\theta_{f^X}$, and $l(\cdot,\cdot)$ measures square error in our experiments. We then use the estimated treatment $\hat{X}$ (in the first stage) to regress the outcome $Y$ in the second stage:
\begin{equation}\label{equ-y}
    \mathcal{L}_{Y}=\frac{1}{N}\sum_{i=1}^Nl(y_i,f^Y(\phi^C(v_i),f^{emb}(f^X(\phi^Z(v_i),\phi^C(v_i))))),
\end{equation}
where $f^{emb}$ is an embedding network with parameter $\theta_{f^{emb}}$ for expanding the dimension of $\hat{X}$, $f^Y$ is the second-stage (outcome) regression network with parameter $\theta_{f^Y}$. $\mathcal{L}_{X_r}$ and $\mathcal{L}_{Y_r}$ are minimized to optimize the parameters of representation, treatment, embedding, and outcome networks to further improve the decomposed representations. 

Note that we assume that the candidate IVs are independent of the unobserved confounders. Based on our regularization term, the decomposed IV representations meet the relevance and exclusion assumptions. Besides, effect homogeneity and monotonicity assumption are often used in the analysis of instrumental variables. Based on the structural equation model, our algorithm models a homogeneity IV to estimate the accurate structural function of the treatment on the outcome \cite{wright1928tariff, goldberger1972structural, wooldridge2002econometric}.

\subsection{Model Optimization}
\begin{algorithm}[t]  
  \caption{AutoIV: Automatic IV Decomposition}  
  \label{alg}  
  \begin{algorithmic}[1]
    \Require  
      Training set $\mathcal{T}={(\boldsymbol{v}_{i},\boldsymbol{x}_{i},\boldsymbol{y}_{i})}_{i=1}^{N_{T}}$;
      variational distribution parameters $\theta_{ZX}$, $\theta_{ZY}$, $\theta_{CX}$, $\theta_{CY}$, and $\theta_{ZC}$;
      IV and confounder representation networks $\phi^{Z}(\cdot;\theta_{\phi^{Z}})$ and $\phi^{C}(\cdot;\theta_{\phi^{C}})$, respectively;
      treatment regression, embedding, and outcome regression networks $f^{X}(\cdot;\theta_{f^{X}})$, $f^{emb}(\cdot;\theta_{f^{emb}})$, and $f^{Y}(\cdot;\theta_{f^{Y}})$, respectively;
      hyperparameters $\alpha$ and $\eta$; training epochs $M$; batchsize $B$.
    \Ensure  
      Well-trained $\phi^{Z}(\cdot;\hat{\theta}_{\phi^{Z}})$ and $\phi^{C}(\cdot;\hat{\theta}_{\phi^{C}})$
    \State Initialize Adam optimizer and all the parameters;
    \For{$epoch=1$ to $M$}
    \State Randomly sample $B$ examples from $\mathcal{T}$;
    \State Update variational distribution parameters $\theta_{ZX}$, $\theta_{ZY}$, $\theta_{CX}$, $\theta_{CY}$, $\theta_{ZC}$ by minimizing $\mathcal{L}^{LLD}$ as Eq. (\ref{equ-lld});
    \State Update representation networks parameters $\theta_{\phi^{Z}}$ and $\theta_{\phi^{C}}$ by minimizing $\mathcal{L}^{MI}$ as Eq. (\ref{equ-mi});
    \State Update representation and treatment regression network parameters $\theta_{\phi^{Z}}$, $\theta_{\phi^{C}}$, $\theta_{f^{X}}$ by minimizing $\mathcal{L}_{X}$ as Eq. (\ref{equ-x});
    \State Update representation, embedding, and outcome regression network parameters $\theta_{\phi^{Z}}$, $\theta_{\phi^{C}}$, $\theta_{f_{emb}}$, $\theta_{f^{Y}}$ by minimizing $\mathcal{L}_{Y}$ as Eq. (\ref{equ-y}).
    \EndFor
  \end{algorithmic}  
\end{algorithm}
As we minimize Eq. (\ref{equ-zx-lld}), (\ref{equ-zy-lld}), (\ref{equ-cx-lld}), (\ref{equ-cy-lld}), and (\ref{equ-zc-lld}) to optimize the parameters $\theta_{ZX}$, $\theta_{ZY}$, $\theta_{CX}$, $\theta_{CY}$, and $\theta_{ZC}$, respectively, each variational distribution approximates the corresponding conditional distribution. We simplify the expression by combining all the variational approximation loss as
\begin{equation}\label{equ-lld}
    \mathcal{L}^{LLD}=\mathcal{L}_{ZX}^{LLD}+\mathcal{L}_{ZY}^{LLD}+\mathcal{L}_{CX}^{LLD}+\mathcal{L}_{CY}^{LLD}+\mathcal{L}_{ZC}^{LLD}.
\end{equation}
Notice that each loss term in Eq. (\ref{equ-lld}) optimizes the corresponding parameters and will not interact with each other. We then combine all the mutual information constraints loss functions of Eq. (\ref{equ-zx-mi}), (\ref{equ-zy-mi}), (\ref{equ-cx-mi}), (\ref{equ-cy-mi}), and (\ref{equ-zc-mi}) as
\begin{equation}\label{equ-mi}
    \mathcal{L}^{MI}=\mathcal{L}_{ZX}^{MI}+\mathcal{L}_{ZY}^{MI}+\alpha(\mathcal{L}_{CX}^{MI}+\mathcal{L}_{CY}^{MI})+\eta\mathcal{L}_{ZC}^{MI},
\end{equation}
where $\alpha$ and $\eta$ are hyper-parameters tuned on a held-out validation set. Eq. (\ref{equ-mi}) is minimized to optimize the representation networks $\phi^Z(\cdot)$ and $\phi^C(\cdot)$ with parameters $\theta_{\phi^Z}$ and $\theta_{\phi^C}$. Eq. (\ref{equ-x}) is minimized to optimize parameters of the representation and treatment networks (i.e., $\theta_{\phi^{Z}}$, $\theta_{\phi^{C}}$, and $\theta_{f^X}$), and Eq. (\ref{equ-y}) is minimized to optimize the parameters of the representation, embedding, and outcome networks (i.e., $\theta_{\phi^{Z}}$, $\theta_{\phi^{C}}$, $\theta_{f^{emb}}$, and $\theta_{f^Y}$). We optimize Eq. (\ref{equ-lld}), (\ref{equ-mi}), (\ref{equ-x}), and (\ref{equ-y}) for the corresponding parameters alternately to get optimal decomposed representations of IVs and confounders. Finally, we use the generated representations for counterfactual prediction with downstream IV-based methods to testify the validity of the learned representations. The whole optimization procedure of our AutoIV algorithm is stated in Algorithm \ref{alg}. 


\section{Experiments}\label{sec-exp}
\begin{figure}[t]
    \centering
    \includegraphics[trim={4.5cm 0.3cm 4.5cm 0.3cm},width=0.65\columnwidth]{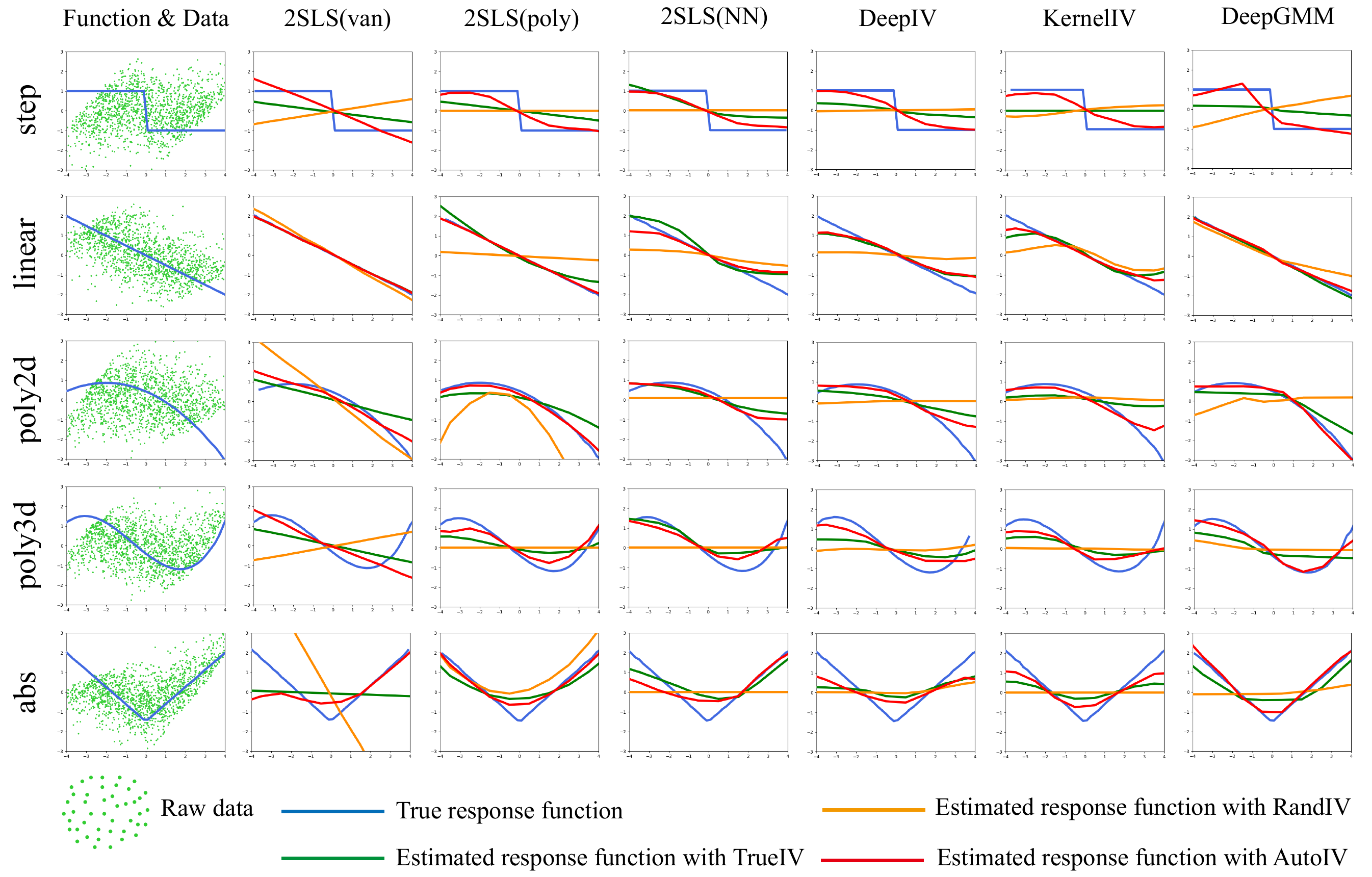}
    \caption{Response function estimation in low-dimensional scenarios.}
    \label{fig-toy}
\end{figure}

\begin{table*}[t]
\caption{Results (MSE$\pm$Std) in low-dimensional scenarios over 20 runs.}
     \label{toy-table}
\centering
\scalebox{0.73}[0.73]{
\renewcommand\tabcolsep{14.0pt}
\begin{tabular}{llcccccc}
\toprule
Methods & IV & step  & abs & linear & poly2d & poly3      \\
\midrule
DirectNN & -  & $2.86\pm0.08$ & $2.38\pm0.05$ & $0.71\pm0.06$ & $2.34\pm0.17$ & $1.71\pm0.07$  \\
\midrule
\multirow{8}{*}{2SLS (van)}
&RandIV           & $2.72\pm0.82$ & $2.61\pm0.40$ & $0.33\pm0.31$ & $1.75\pm0.80$ & $1.69\pm0.77$\\ 
&TrueIV           & $0.77\pm0.07$ & $2.05\pm0.10$ & $0.09\pm0.06$ & $0.40\pm0.04$ & $0.25\pm0.06$ \\
&UAS (w/o $Z$)    & $3.41\pm0.17$ & $3.59\pm0.17$ & $0.95\pm0.10$ & $3.02\pm0.23$ & $2.34\pm0.07$\\
&UAS (w/ $Z$)     & $2.15\pm0.14$ & $2.76\pm0.12$ & $0.28\pm0.04$ & $1.76\pm0.14$ & $1.25\pm0.08$\\
&WAS (w/o $Z$)    & $3.39\pm0.16$ & $3.58\pm0.17$ & $0.94\pm0.10$ & $3.00\pm0.23$ & $2.32\pm0.07$\\
&WAS (w/ $Z$)     & $2.10\pm0.18$ & $2.76\pm0.08$ & $0.28\pm0.09$ & $1.72\pm0.18$ & $1.25\pm0.10$\\
&AutoIV (w/o $Z$) & $1.20\pm1.96$ & $\boldsymbol{1.11\pm0.09}$ & $\boldsymbol{0.00\pm0.00}$ & $0.22\pm0.13$ & $0.21\pm0.06$ \\
&AutoIV (w/ $Z$)  & $\boldsymbol{0.38\pm0.03}$ & $1.41\pm0.87$ & $\boldsymbol{0.00\pm0.00}$ & $\boldsymbol{0.21\pm0.08}$ & $\boldsymbol{0.19\pm0.04}$\\
\midrule
\multirow{8}{*}{2SLS (poly)}
&RandIV           & $2.18\pm0.70$ & $2.21\pm0.38$ & $0.87\pm0.15$ & $1.85\pm0.71$ & $1.28\pm0.28$\\ 
&TrueIV           & $0.86\pm0.16$ & $1.96\pm0.13$ & $0.11\pm0.07$ & $0.43\pm0.07$ & $0.28\pm0.06$ \\
&UAS (w/o $Z$)    & $3.37\pm0.14$ & $3.48\pm0.28$ & $0.98\pm0.06$ & $3.00\pm0.18$ & $2.32\pm0.13$\\
&UAS (w/ $Z$)     & $2.11\pm0.17$ & $2.51\pm0.31$ & $0.28\pm0.04$ & $1.63\pm0.12$ & $1.23\pm0.09$\\
&WAS (w/o $Z$)    & $3.34\pm0.15$ & $3.55\pm0.28$ & $0.98\pm0.06$ & $2.92\pm0.22$ & $2.30\pm0.11$\\
&WAS (w/ $Z$)     & $2.06\pm0.19$ & $2.50\pm0.28$ & $0.28\pm0.10$ & $1.59\pm0.17$ & $1.26\pm0.09$\\
&AutoIV (w/o $Z$) & $\boldsymbol{0.39\pm0.02}$ & $0.41\pm0.11$ & $\boldsymbol{0.00\pm0.00}$ & $\boldsymbol{0.17\pm0.08}$ & $0.19\pm0.05$ \\
&AutoIV (w/ $Z$)  & $\boldsymbol{0.39\pm0.02}$ & $\boldsymbol{0.28\pm0.04}$ & $\boldsymbol{0.00\pm0.00}$ & $0.28\pm0.21$ & $\boldsymbol{0.18\pm0.04}$\\
\midrule
\multirow{8}{*}{2SLS (NN)}
&RandIV           & $1.26\pm0.04$ & $2.09\pm0.26$ & $0.97\pm0.05$ & $0.99\pm0.07$ & $1.02\pm0.02$\\ 
&TrueIV           & $1.04\pm0.12$ & $1.99\pm0.20$ & $\boldsymbol{0.14\pm0.02}$ & $0.58\pm0.06$ & $0.32\pm0.08$ \\
&UAS (w/o $Z$)    & $2.46\pm0.09$ & $3.33\pm0.32$ & $0.97\pm0.05$ & $2.19\pm0.15$ & $2.08\pm0.06$\\
&UAS (w/ $Z$)     & $1.26\pm0.04$ & $2.12\pm0.34$ & $0.97\pm0.05$ & $0.99\pm0.07$ & $1.02\pm0.03$\\
&WAS (w/o $Z$)    & $2.45\pm0.06$ & $3.42\pm0.31$ & $0.97\pm0.05$ & $2.20\pm0.12$ & $2.13\pm0.10$\\
&WAS (w/ $Z$)     & $1.82\pm0.17$ & $2.68\pm0.20$ & $0.36\pm0.09$ & $1.61\pm0.15$ & $1.17\pm0.10$\\
&AutoIV (w/o $Z$) & $0.47\pm0.17$ & $0.50\pm0.18$ & $0.30\pm0.23$ & $0.50\pm0.16$ & $0.33\pm0.16$ \\
&AutoIV (w/ $Z$)  & $\boldsymbol{0.37\pm0.09}$ & $\boldsymbol{0.35\pm0.06}$ & $0.25\pm0.09$ & $\boldsymbol{0.45\pm0.30}$ & $\boldsymbol{0.26\pm0.14}$\\
\midrule
\multirow{8}{*}{DeepIV}
&RandIV           & $1.50\pm0.09$ & $1.76\pm0.33$ & $0.90\pm0.05$ & $1.41\pm0.11$ & $1.15\pm0.11$\\ 
&TrueIV           & $1.34\pm0.09$ & $1.69\pm0.26$ & $0.72\pm0.05$ & $1.33\pm0.12$ & $1.01\pm0.09$ \\
&UAS (w/o $Z$)    & $1.64\pm0.11$ & $1.95\pm0.30$ & $0.93\pm0.06$ & $1.53\pm0.21$ & $1.28\pm0.13$\\
&UAS (w/ $Z$)     & $1.59\pm0.08$ & $1.82\pm0.24$ & $0.71\pm0.06$ & $1.37\pm0.13$ & $1.13\pm0.07$\\
&WAS (w/o $Z$)    & $1.77\pm0.16$ & $1.81\pm0.29$ & $0.94\pm0.07$ & $1.49\pm0.15$ & $1.34\pm0.12$\\
&WAS (w/ $Z$)     & $1.59\pm0.12$ & $1.76\pm0.30$ & $0.70\pm0.07$ & $1.38\pm0.12$ & $1.08\pm0.09$\\
&AutoIV (w/o $Z$) & $\boldsymbol{0.66\pm0.16}$ & $0.90\pm0.13$ & $0.70\pm0.18$ & $0.80\pm0.14$ & $0.86\pm0.12$ \\
&AutoIV (w/ $Z$)  & $0.72\pm0.17$ & $\boldsymbol{0.86\pm0.08}$ & $\boldsymbol{0.63\pm0.11}$ & $\boldsymbol{0.71\pm0.20}$ & $\boldsymbol{0.67\pm0.13}$\\
\midrule
\multirow{8}{*}{KernelIV}
&RandIV           & $1.55\pm0.17$ & $4.79\pm0.13$ & $0.94\pm0.11$ & $1.04\pm0.02$ & $1.10\pm0.15$\\ 
&TrueIV           & $1.24\pm0.11$ & $3.67\pm0.68$ & $\boldsymbol{0.67\pm0.06}$ & $1.01\pm0.02$ & $0.99\pm0.03$ \\
&UAS (w/o $Z$)    & $3.15\pm0.28$ & $5.42\pm0.13$ & $0.92\pm0.11$ & $2.41\pm0.14$ & $1.86\pm0.09$\\
&UAS (w/ $Z$)     & $2.37\pm0.32$ & $5.04\pm0.21$ & $\boldsymbol{0.67\pm0.06}$ & $1.77\pm0.07$ & $1.22\pm0.38$\\
&WAS (w/o $Z$)    & $3.22\pm0.44$ & $5.39\pm0.16$ & $0.94\pm0.11$ & $2.48\pm0.26$ & $1.88\pm0.09$\\
&WAS (w/ $Z$)     & $2.40\pm0.24$ & $4.81\pm0.40$ & $\boldsymbol{0.67\pm0.06}$ & $1.77\pm0.09$ & $1.04\pm0.02$\\
&AutoIV (w/o $Z$) & $0.93\pm0.05$ & $1.07\pm0.05$ & $0.92\pm0.08$ & $1.03\pm0.09$ & $0.90\pm0.36$ \\
&AutoIV (w/ $Z$)  & $\boldsymbol{0.80\pm0.17}$ & $\boldsymbol{0.90\pm0.04}$ & $0.78\pm0.09$ & $\boldsymbol{0.78\pm0.27}$ & $\boldsymbol{0.89\pm0.10}$\\
\midrule
\multirow{8}{*}{DeepGMM}
&RandIV           & $2.03\pm0.62$ & $2.53\pm0.31$ & $0.86\pm0.21$ & $2.16\pm0.48$ & $1.35\pm0.43$\\ 
&TrueIV           & $1.03\pm0.10$ & $1.69\pm0.28$ & $\boldsymbol{0.12\pm0.05}$ & $0.48\pm0.08$ & $0.32\pm0.14$ \\
&UAS (w/o $Z$)    & $3.65\pm0.13$ & $2.23\pm0.76$ & $1.00\pm0.07$ & $3.53\pm0.65$ & $2.32\pm0.32$\\
&UAS (w/ $Z$)     & $2.23\pm0.07$ & $2.33\pm0.60$ & $0.47\pm0.04$ & $1.75\pm0.19$ & $1.12\pm0.14$\\
&WAS (w/o $Z$)    & $3.74\pm0.14$ & $2.21\pm0.81$ & $1.01\pm0.07$ & $3.47\pm0.50$ & $2.31\pm0.31$\\
&WAS (w/ $Z$)     & $2.22\pm0.18$ & $2.33\pm0.59$ & $0.43\pm0.10$ & $1.75\pm0.18$ & $1.13\pm0.15$\\
&AutoIV (w/o $Z$) & $0.71\pm0.36$ & $0.66\pm0.58$ & $0.44\pm0.29$ & $0.38\pm0.48$ & $0.37\pm0.36$ \\
&AutoIV (w/ $Z$)  & $\boldsymbol{0.69\pm0.43}$ & $\boldsymbol{0.35\pm0.42}$ & $0.28\pm0.30$ & $\boldsymbol{0.22\pm0.17}$ & $\boldsymbol{0.24\pm0.17}$\\
\bottomrule
\end{tabular}}
\end{table*}

\begin{table}[t]
\caption{Ablation experiments of AutoIV.}
    \label{table-ablation}
\scalebox{0.82}[0.82]{
\renewcommand\tabcolsep{14.0pt}
\begin{tabular}{lccccc}
\toprule
Methods  & $\mathcal{L}_{ZX_{m}}+\mathcal{L}_{ZY_{m}}$ & $\mathcal{L}_{CX_{m}}+\mathcal{L}_{CY_{m}}$ & $\mathcal{L}_{CZ_{m}}$ & $\mathcal{L}_{X_{r}}+\mathcal{L}_{Y_{r}}$ & Results \\
\midrule
\multirow{5}{*}{DeepIV}   &                                             & $\checkmark$                                & $\checkmark$           & $\checkmark$                              & $0.95\pm0.05$   \\
         & $\checkmark$                                &                                             & $\checkmark$           & $\checkmark$                              & $0.92\pm0.06$   \\
         & $\checkmark$                                & $\checkmark$                                &                        & $\checkmark$                              & $0.96\pm0.06$   \\
         & $\checkmark$                                & $\checkmark$                                & $\checkmark$           &                                           & $0.98\pm0.06$   \\
         & $\checkmark$                                & $\checkmark$                                & $\checkmark$           & $\checkmark$                              & $\boldsymbol{0.86\pm0.08}$   \\
\midrule
\multirow{5}{*}{KernelIV} &                                             & $\checkmark$                                & $\checkmark$           & 
$\checkmark$                              & $0.91\pm0.09$   \\
         & $\checkmark$                                &                                             & $\checkmark$           & $\checkmark$                              & $0.98\pm0.11$   \\
         & $\checkmark$                                & $\checkmark$                                &                        & $\checkmark$                              & $1.11\pm0.15$   \\
         & $\checkmark$                                & $\checkmark$                                & $\checkmark$           &                                           & $>10$   \\
         & $\checkmark$                                & $\checkmark$                                & $\checkmark$           & $\checkmark$                              & $\boldsymbol{0.90\pm0.04}$  \\
\midrule
\multirow{5}{*}{DeepGMM}  &                                             & $\checkmark$                                & $\checkmark$           & 
$\checkmark$                              & $0.49\pm0.25$   \\
         & $\checkmark$                                &                                             & $\checkmark$           & $\checkmark$                              & $0.60\pm0.62$   \\
         & $\checkmark$                                & $\checkmark$                                &                        & $\checkmark$                              & $0.68\pm0.31$   \\
         & $\checkmark$                                & $\checkmark$                                & $\checkmark$           &                                           & $0.73\pm0.57$   \\
         & $\checkmark$                                & $\checkmark$                                & $\checkmark$           & $\checkmark$                              & $\boldsymbol{0.35\pm0.42}$   \\
\bottomrule
\end{tabular}}
\end{table}

\begin{figure}[t]
    \centering
    \includegraphics[trim={4cm 0cm 4cm 0cm},width=1.0\columnwidth]{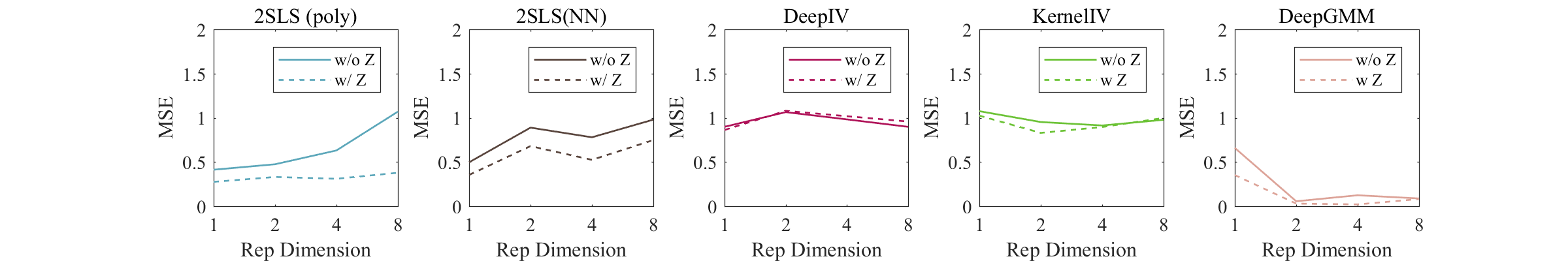} 
    \caption{Performance of AutoIV by varying representation dimensions.}
    \label{fig-rep_dim}
\end{figure}

\begin{figure*}[t]
    \centering
    \includegraphics[trim={6cm 0cm 10cm 0cm},width=1.0\columnwidth]{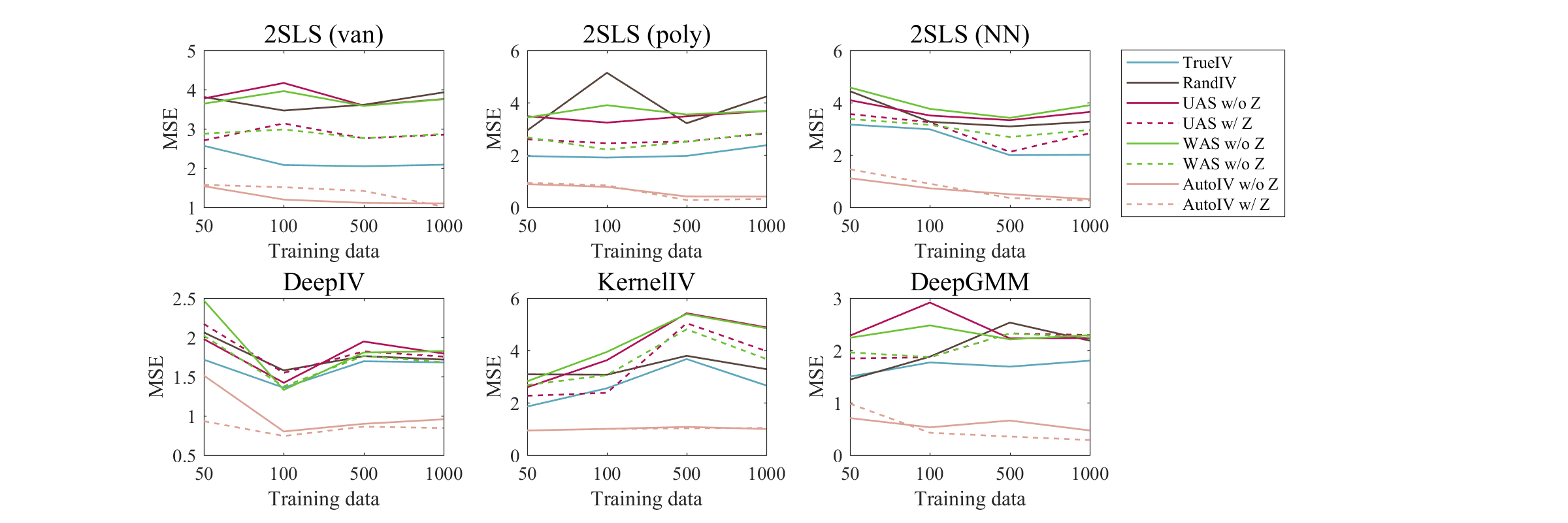}
    \caption{Performance of AutoIV by varying the training data size.}
    \label{fig-num}
\end{figure*}

\begin{figure}[t]
    \centering
    \includegraphics[trim={3cm 0cm 9cm 0cm},width=1.0\columnwidth]{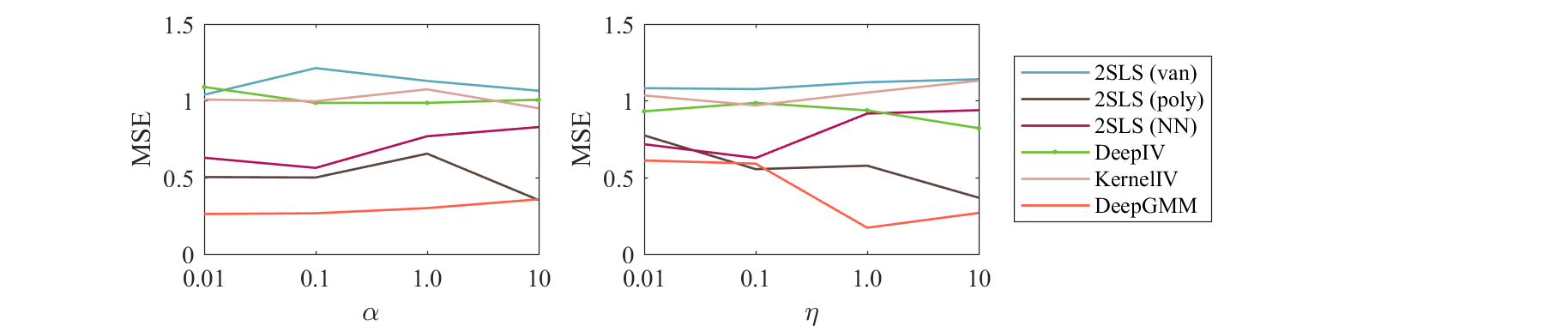}
    \caption{Results of sensitivity analysis of the hyperparameters $\alpha$ (left) and $\eta$ (right) in the AutoIV algorithm.}
    \label{fig-hyperparameter}
\end{figure}

\begin{table*}[t]
    \caption{Results of high-dimensional experiments on MNIST data with representation dimension as 5, 10, and 15.}
    \label{table-mnist}
    \scalebox{0.8}[0.8]{
    \renewcommand\tabcolsep{9.0pt}
    \begin{tabular}{llccccc}
    \toprule
    Methods    & Scenarios           & RandIV           & TrueIV                         & AutoIV-5        & AutoIV-10       & AutoIV-15       \\
    \midrule
    \multirow{3}{*}{2SLS(van)}  & ${\rm{MNIST}}_{Z}$  & $1.688\pm0.229$  & $\boldsymbol{0.986\pm0.030}$                & $0.994\pm0.045$ & $1.046\pm0.055$ & $0.995\pm0.042$ \\
    & ${\rm{MNIST}}_{C}$  & $1.657\pm0.170$  & $1.022\pm0.046$                & $\boldsymbol{0.999\pm0.040}$ & $1.031\pm0.053$ & $1.018\pm0.047$ \\
    & ${\rm{MNIST}}_{ZC}$ & $2.053\pm0.307$  & $1.780\pm0.283$                & $1.006\pm0.041$ & $1.019\pm0.040$ & $\boldsymbol{0.999\pm0.033}$ \\
    \midrule
    \multirow{3}{*}{2SLS(poly)} & ${\rm{MNIST}}_{Z}$  & $1.792\pm1.411$  & $0.977\pm0.032$                & $\boldsymbol{0.444\pm0.142}$ & $0.604\pm0.304$ & $0.530\pm0.207$ \\
    & ${\rm{MNIST}}_{C}$  & $1.491\pm1.307$  & $0.982\pm0.041$                & $\boldsymbol{0.426\pm0.053}$ & $0.533\pm0.207$ & $0.676\pm0.306$ \\
    & ${\rm{MNIST}}_{ZC}$ & $1.327\pm0.570$  & $1.001\pm0.027$                & $\boldsymbol{0.703\pm0.219}$ & $0.928\pm0.065$ & $0.976\pm0.036$ \\
    \midrule
    \multirow{3}{*}{2SLS(NN)}   & ${\rm{MNIST}}_{Z}$  & $1.382\pm0.110$  & $1.045\pm0.068$                & $0.663\pm0.219$ & $0.369\pm0.066$ & $\boldsymbol{0.336\pm0.043}$ \\
    & ${\rm{MNIST}}_{C}$  & $1.352\pm0.073$  & $1.074\pm0.077$                & $0.785\pm0.196$ & $0.374\pm0.072$ & $\boldsymbol{0.323\pm0.046}$ \\
    & ${\rm{MNIST}}_{ZC}$ & $1.501\pm0.068$  & $1.427\pm0.076$                & $0.967\pm0.081$ & $0.881\pm0.142$ & $\boldsymbol{0.829\pm0.224}$ \\
    \midrule
    \multirow{3}{*}{DeepIV}     & ${\rm{MNIST}}_{Z}$  & $1.102\pm0.0912$ & $1.030\pm0.054$                & $\boldsymbol{0.875\pm0.135}$ & $0.891\pm0.053$ & $0.985\pm0.117$ \\
    & ${\rm{MNIST}}_{C}$  & $1.221\pm0.107$  & $1.590\pm0.402$                & $\boldsymbol{0.956\pm0.118}$ & $1.111\pm0.144$ & $1.191\pm0.088$ \\
    & ${\rm{MNIST}}_{ZC}$ & $1.163\pm0.240$  & $1.269\pm0.336$                & $\boldsymbol{1.047\pm0.033}$ & $1.191\pm0.119$ & $1.088\pm0.106$ \\
    \midrule
    \multirow{3}{*}{KernelIV}   & ${\rm{MNIST}}_{Z}$  & $0.978\pm0.034$  & $0.984\pm0.038$                & $0.968\pm0.037$ & $0.967\pm0.034$ & $\boldsymbol{0.941\pm0.044}$ \\
    & ${\rm{MNIST}}_{C}$  & $0.979\pm0.038$  & $0.979\pm0.038$                & $\boldsymbol{0.960\pm0.033}$ & $0.972\pm0.037$ & $0.977\pm0.034$ \\
    & ${\rm{MNIST}}_{ZC}$ & $0.984\pm0.034$  & $0.984\pm0.034$                & $\boldsymbol{0.944\pm0.052}$ & $0.966\pm0.036$ & $0.966\pm0.036$ \\
    \midrule
    \multirow{3}{*}{DeepGMM}    & ${\rm{MNIST}}_{Z}$  & $1.040\pm0.213$  & $0.586\pm0.225$                & $0.229\pm0.333$ & $\boldsymbol{0.064\pm0.091}$ & $0.124\pm0.227$ \\
    & ${\rm{MNIST}}_{X}$  & $1.108\pm0.255$  & $0.923\pm0.086$                & $\boldsymbol{0.122\pm0.182}$ & $0.204\pm0.309$ & $0.495\pm0.394$ \\
    & ${\rm{MNIST}}_{ZC}$ & $1.051\pm0.242$  & $0.471\pm0.129$                & $0.026\pm0.019$ & $\boldsymbol{0.012\pm0.009}$ & $0.014\pm0.014$ \\
    \bottomrule
    \end{tabular}}
\end{table*}

\begin{figure}[t]
    \centering
    \includegraphics[trim={-6cm 0cm -6cm 0cm},width=0.3\columnwidth]{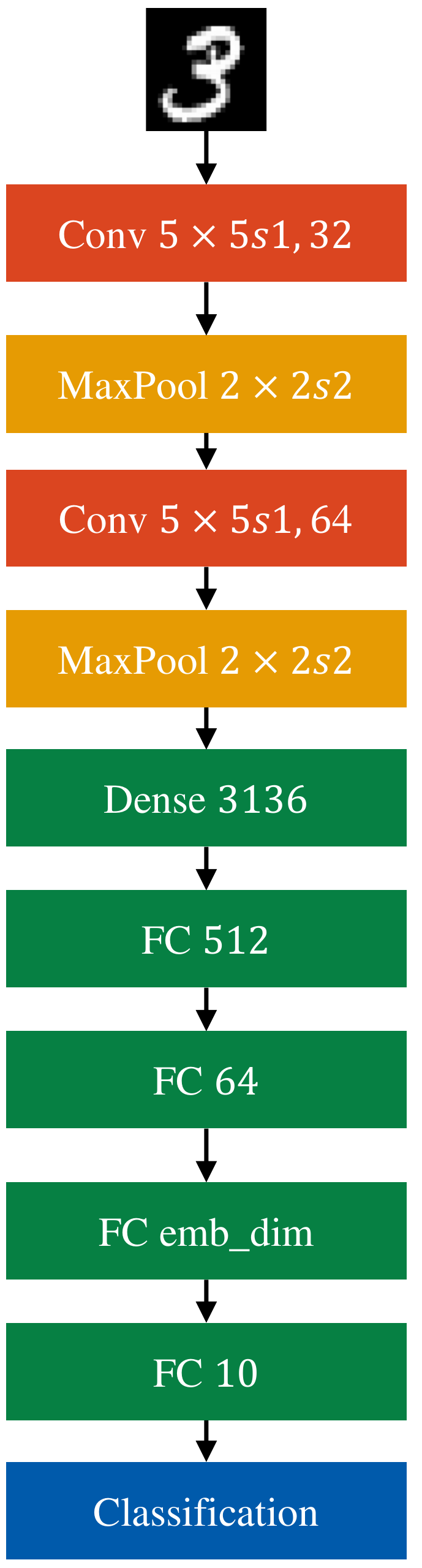}
    \caption{Convolutional networks for MNIST data. The data are sampled on the penultimate fully-connected (FC) layer.}
    \label{fig-conv}
\end{figure}

\begin{table}[t]
    \caption{Results of high-dimensional scenarios on MNIST data with different data composition.}
    \label{table-mnist-data}
    \renewcommand\tabcolsep{37.0pt}
    \scalebox{0.6}[0.6]{
    \begin{tabular}{llccc}
    \toprule
        \multicolumn{5}{c}{$d_{Z}$: 5, $d_{F}$: 10, $d_{A}$: 4, $d_{U}$: 1}             \\
        \hline\hline
        Methods                                           & Scenarios           & RandIV              & TrueIV              & AutoIV              \\
        \hline
        \multirow{3}{*}{DirectNN}                         & ${\rm{MNIST}}_{Z}$  & $1.0314 \pm 0.0584$ & -                   & -                   \\
                                                      & ${\rm{MNIST}}_{C}$  & $1.3136 \pm 0.1213$ & -                   & -                   \\
                                                      & ${\rm{MNIST}}_{ZC}$ & $1.5746 \pm 0.1054$ & -                   & -                   \\
                                                      \hline\multirow{3}{*}{2SLS(van)}                        & ${\rm{MNIST}}_{Z}$  & $1.8015 \pm 0.4056$ & $\boldsymbol{0.9827 \pm 0.0403}$ & $0.9897 \pm 0.0469$ \\
                                                      & ${\rm{MNIST}}_{C}$  & $1.6923 \pm 0.3306$ & $\boldsymbol{0.9880 \pm 0.0417}$ & $1.0100 \pm 0.0820$ \\
                                                      & ${\rm{MNIST}}_{ZC}$ & $1.8393 \pm 0.3410$ & $1.4693 \pm 0.1897$ & $\boldsymbol{1.0064 \pm 0.0296}$ \\
                                                      \hline\multirow{3}{*}{2SLS(poly)}                       & ${\rm{MNIST}}_{Z}$  & $0.9996 \pm 0.0573$ & $0.9783 \pm 0.0378$ & $\boldsymbol{0.4085 \pm 0.1414}$ \\
                                                      & ${\rm{MNIST}}_{C}$  & $0.9757 \pm 0.0446$ & $0.9666 \pm 0.0397$ & $\boldsymbol{0.4965 \pm 0.1397}$ \\
                                                      & ${\rm{MNIST}}_{ZC}$ & $0.9815 \pm 0.0325$ & $0.9816 \pm 0.0324$ & $\boldsymbol{0.8666 \pm 0.1465}$ \\
                                                      \hline\multirow{3}{*}{2SLS(NN)}                         & ${\rm{MNIST}}_{Z}$  & $1.1456 \pm 0.1052$ & $0.7765 \pm 0.0358$ & $\boldsymbol{0.2601 \pm 0.0273}$ \\
                                                      & ${\rm{MNIST}}_{C}$  & $1.3150 \pm 0.1590$ & $0.8302 \pm 0.0693$ & $\boldsymbol{0.3698 \pm 0.0473}$ \\
                                                      & ${\rm{MNIST}}_{ZC}$ & $1.2535 \pm 0.0574$ & $1.1636 \pm 0.0886$ & $\boldsymbol{0.7349 \pm 0.2369}$ \\
                                                      \hline\multirow{3}{*}{DeepIV}                           & ${\rm{MNIST}}_{Z}$  & $1.0356 \pm 0.1509$ & $1.2036 \pm 0.2065$ & $\boldsymbol{0.8174 \pm 0.1286}$ \\
                                                      & ${\rm{MNIST}}_{C}$  & $1.1665 \pm 0.2003$ & $\boldsymbol{1.0873 \pm 0.1486}$ & $1.1090 \pm 0.1750$ \\
                                                      & ${\rm{MNIST}}_{ZC}$ & $1.5355 \pm 0.2431$ & $1.2730 \pm 0.2021$ & $\boldsymbol{1.0168 \pm 0.0727}$ \\
                                                      \hline\multirow{3}{*}{KernelIV}                         & ${\rm{MNIST}}_{Z}$  & $0.9791 \pm 0.0392$ & $0.9827 \pm 0.0374$ & $\boldsymbol{0.9583 \pm 0.0477}$ \\
                                                      & ${\rm{MNIST}}_{C}$  & $0.9671 \pm 0.0370$ & $0.9671 \pm 0.0370$ & $\boldsymbol{0.9604 \pm 0.0408}$ \\
                                                      & ${\rm{MNIST}}_{ZC}$ & $0.9826 \pm 0.0339$ & $0.9826 \pm 0.0339$ & $\boldsymbol{0.9673 \pm 0.0351}$ \\
                                                      \hline\multirow{3}{*}{DeepGMM}                          & ${\rm{MNIST}}_{Z}$  & $0.9679 \pm 0.1546$ & $0.5502 \pm 0.2183$ & $\boldsymbol{0.3052 \pm 0.3722}$ \\
                                                      & ${\rm{MNIST}}_{C}$  & $1.1639 \pm 0.2048$ & $0.9097 \pm 0.0944$ & $\boldsymbol{0.1827 \pm 0.2632}$ \\
                                                      & ${\rm{MNIST}}_{ZC}$ & $1.0206 \pm 0.1458$ & $0.4286 \pm 0.0566$ & $\boldsymbol{0.0074 \pm 0.0058}$ \\
                                                      \hline\hline\multicolumn{5}{c}{$d_{Z}$: 10, $d_{F}$: 5, $d_{A}$: 4, $d_{U}$: 1}                     \\
    \hline\hline Methods                                           & Scenarios           & RandIV              & TrueIV              & AutoIV              \\
    \hline\multirow{3}{*}{DirectNN}                         & ${\rm{MNIST}}_{Z}$  & $1.3030 \pm 0.0706$ & -                   & -                   \\
                                                      & ${\rm{MNIST}}_{C}$  & $1.2205 \pm 0.0799$ & -                   & -                   \\
                                                      & ${\rm{MNIST}}_{ZC}$ & $1.8299 \pm 0.1313$ & -                   & -                   \\
                                                      \hline\multirow{3}{*}{2SLS(van)}                        & ${\rm{MNIST}}_{Z}$  & $2.0021 \pm 0.1951$ & $\boldsymbol{0.9894 \pm 0.0282}$ & $1.0096 \pm 0.0515$ \\
                                                      & ${\rm{MNIST}}_{C}$  & $1.7111 \pm 0.1561$ & $\boldsymbol{0.9826 \pm 0.0474}$ & $1.0242 \pm 0.0331$ \\
                                                      & ${\rm{MNIST}}_{ZC}$ & $1.9091 \pm 0.2820$ & $0.9990 \pm 0.0394$ & $\boldsymbol{0.9981 \pm 0.0543}$ \\
                                                      \hline\multirow{3}{*}{2SLS(poly)}                       & ${\rm{MNIST}}_{Z}$  & $2.5048 \pm 2.3093$ & $0.9878 \pm 0.0298$ & $\boldsymbol{0.6709 \pm 0.2765}$ \\
                                                      & ${\rm{MNIST}}_{C}$  & $2.0190 \pm 1.5183$ & $0.9761 \pm 0.0440$ & $\boldsymbol{0.3794 \pm 0.0876}$ \\
                                                      & ${\rm{MNIST}}_{ZC}$ & $1.6373 \pm 0.6966$ & $0.9692 \pm 0.0310$ & $\boldsymbol{0.8461 \pm 0.1283}$ \\
                                                      \hline\multirow{3}{*}{2SLS(NN)}                         & ${\rm{MNIST}}_{Z}$  & $1.3283 \pm 0.1149$ & $0.8832 \pm 0.0789$ & $\boldsymbol{0.4526 \pm 0.1766}$ \\
                                                      & ${\rm{MNIST}}_{C}$  & $1.1980 \pm 0.0629$ & $0.9304 \pm 0.0505$ & $\boldsymbol{0.3484 \pm 0.0229}$ \\
                                                      & ${\rm{MNIST}}_{ZC}$ & $1.3864 \pm 0.0744$ & $1.0940 \pm 0.0645$ & $\boldsymbol{0.6992 \pm0.1224}$  \\
                                                      \hline\multirow{3}{*}{DeepIV}                           & ${\rm{MNIST}}_{Z}$  & $1.1037 \pm 0.1253$ & $1.1432 \pm 0.1797$ & $\boldsymbol{0.9850 \pm 0.1336}$ \\
                                                      & ${\rm{MNIST}}_{C}$  & $1.2980 \pm 0.0987$ & $1.1619 \pm 0.2628$ & $\boldsymbol{1.0055 \pm 0.1111}$ \\
                                                      & ${\rm{MNIST}}_{ZC}$ & $1.1620 \pm 0.2386$ & $1.5090 \pm 0.4130$ & $\boldsymbol{0.9963 \pm 0.0733}$ \\
                                                      \hline\multirow{3}{*}{KernelIV}                         & ${\rm{MNIST}}_{Z}$  & $0.9815 \pm 0.0219$ & $0.9839 \pm 0.0276$ & $\boldsymbol{0.9404 \pm 0.0547}$ \\
                                                      & ${\rm{MNIST}}_{C}$  & $0.9771 \pm 0.0427$ & $0.9771 \pm 0.0427$ & $\boldsymbol{0.9613 \pm 0.0379}$ \\
                                                      & ${\rm{MNIST}}_{ZC}$ & $0.9769 \pm 0.0326$ & $0.9769 \pm 0.0326$ & $\boldsymbol{0.9523 \pm 0.0335}$ \\
                                                      \hline\multirow{3}{*}{DeepGMM}                          & ${\rm{MNIST}}_{Z}$  & $1.0743 \pm 0.1008$ & $0.7820 \pm 0.2455$ & $\boldsymbol{0.0367 \pm 0.0535}$ \\
                                                      & ${\rm{MNIST}}_{C}$  & $1.0260 \pm 0.1733$ & $0.9021 \pm 0.1084$ & $\boldsymbol{0.2259 \pm 0.2791}$ \\
                                                      & ${\rm{MNIST}}_{ZC}$ & $1.1627 \pm 0.2468$ & $0.4830 \pm 0.1252$ & $\boldsymbol{0.0107 \pm 0.0082}$ \\
                                                      \hline\hline\multicolumn{5}{c}{$d_{Z}$: 10, $d_{F}$: 10, $d_{A}$: 4, $d_{U}$: 1}                 \\
    \hline\hline
    Methods                                           & Scenarios           & RandIV              & TrueIV              & AutoIV              \\
    \hline\multirow{3}{*}{DirectNN}                         & ${\rm{MNIST}}_{Z}$  & $1.3170 \pm 0.0568$ & -                   & -                   \\
                                                      & ${\rm{MNIST}}_{C}$  & $1.3577 \pm 0.0742$ & -                   & -                   \\
                                                      & ${\rm{MNIST}}_{ZC}$ & $1.6820 \pm 0.1113$ & -                   & -                   \\
                                                      \hline\multirow{3}{*}{2SLS(van)}                        & ${\rm{MNIST}}_{Z}$  & $1.6878 \pm 0.2286$ & $\boldsymbol{0.9861 \pm 0.0297}$ & $1.0464 \pm 0.0552$ \\
                                                      & ${\rm{MNIST}}_{C}$  & $1.6570 \pm 0.1699$ & $\boldsymbol{1.0223 \pm 0.0460}$ & $1.0312 \pm 0.0527$ \\
                                                      & ${\rm{MNIST}}_{ZC}$ & $2.0527 \pm 0.3070$ & $1.7801 \pm 0.2825$ & $\boldsymbol{1.0192 \pm 0.0399}$ \\
                                                      \hline\multirow{3}{*}{2SLS(poly)}                       & ${\rm{MNIST}}_{Z}$  & $1.7924 \pm 1.4106$ & $0.9771 \pm 0.0315$ & $\boldsymbol{0.6043 \pm 0.3041}$ \\
                                                      & ${\rm{MNIST}}_{C}$  & $1.4912 \pm 1.3071$ & $0.9824 \pm 0.0414$ & $\boldsymbol{0.5331 \pm 0.2065}$ \\
                                                      & ${\rm{MNIST}}_{ZC}$ & $1.3272 \pm 0.5701$ & $1.0008 \pm 0.0267$ & $\boldsymbol{0.9277 \pm 0.0652}$ \\
                                                      \hline\multirow{3}{*}{2SLS(NN)}                         & ${\rm{MNIST}}_{Z}$  & $1.3819 \pm 0.1103$ & $1.0454 \pm 0.0675$ & $\boldsymbol{0.3687 \pm 0.0656}$ \\
                                                      & ${\rm{MNIST}}_{C}$  & $1.3521 \pm 0.0727$ & $1.0743 \pm0.0768$  & $\boldsymbol{0.3740 \pm 0.0722}$ \\
                                                      & ${\rm{MNIST}}_{ZC}$ & $1.5010 \pm 0.0678$ & $1.4271 \pm 0.0759$ & $\boldsymbol{0.8810 \pm 0.1422}$ \\
                                                      \hline\multirow{3}{*}{DeepIV}                           & ${\rm{MNIST}}_{Z}$  & $1.1016 \pm 0.0912$ & $1.0304 \pm 0.0535$ & $\boldsymbol{0.8910 \pm 0.0529}$ \\
                                                      & ${\rm{MNIST}}_{C}$  & $1.2205 \pm 0.1069$ & $1.5897 \pm 0.4022$ & $\boldsymbol{1.1111 \pm 0.1439}$ \\
                                                      & ${\rm{MNIST}}_{ZC}$ & $1.2625 \pm 0.2401$ & $1.2693 \pm 0.3363$ & $\boldsymbol{1.1914 \pm 0.1190}$ \\
                                                      \hline\multirow{3}{*}{KernelIV}                         & ${\rm{MNIST}}_{Z}$  & $0.9777 \pm 0.0336$ & $0.9838 \pm 0.0383$ & $\boldsymbol{0.9668 \pm 0.0341}$ \\
                                                      & ${\rm{MNIST}}_{C}$  & $0.9793 \pm 0.0377$ & $0.9793 \pm 0.0377$ & $\boldsymbol{0.9724 \pm 0.0368}$ \\
                                                      & ${\rm{MNIST}}_{ZC}$ & $0.9842 \pm 0.0339$ & $0.9842 \pm 0.0339$ & $\boldsymbol{0.9658 \pm 0.0358}$ \\
                                                      \hline\multirow{3}{*}{DeepGMM}                          & ${\rm{MNIST}}_{Z}$  & $1.0401 \pm 0.2125$ & $0.5864 \pm 0.2247$ & $\boldsymbol{0.0640 \pm 0.0906}$ \\
                                                      & ${\rm{MNIST}}_{C}$  & $1.1075 \pm 0.2546$ & $0.9226 \pm 0.0857$ & $\boldsymbol{0.2038 \pm 0.3093}$ \\
                                                      & ${\rm{MNIST}}_{ZC}$ & $1.0513 \pm 0.2420$ & $0.4711 \pm 0.1292$ & $\boldsymbol{0.0123 \pm 0.0094}$\\
                                                      \bottomrule
    \end{tabular}}
\end{table}

In this section, we show the empirical evaluation of applying AutoIV to different downstream IV-based methods for counterfactual prediction. 
The validity of the learned IV representations determines the accuracy of counterfactual prediction of the downstream methods. 
We implement the experiments with Python on a device with CPU Intel Xeon Gold 6254, GPU Nvidia RTX 2080TI, and memory 64MB.

We list the representative IV-based methods introduced previously and used in our experiments in the following.
\begin{itemize}
    \item[1.] \textbf{DirectNN}: directly regress the outcome on the treatment with neural networks. It does not use any information of the IVs, and can be considered as the general supervised learning.
    \item[2.] \textbf{2SLS (van)}: vanilla two-stage least squares with linear models. 
    \item[3.] \textbf{2SLS (poly)}: two-stage least squares with polynomial basis and ridge regularization. 
    \item[4.] \textbf{2SLS (NN)}: two-stage regression with neural networks structure. 
    \item[5.] \textbf{DeepIV} \cite{jason2017deepiv}: fit the treatment with the IVs via optimizing a mixture density network in the first stage, and then fit the outcome by sampling from the mixture density network. We use its original implementation \footnote{https://github.com/jhartford/DeepIV}. 
    \item[6.] \textbf{KernelIV}\cite{singh2019kernel}: a recent kernel method that performs two-stage procedure in reproduce kernel Hilbert spaces. We implement it with Python by referring its original MATLAB version \footnote{https://github.com/r4hu1-5in9h/KernelIV}. 
    The results of ours and original MATLAB version are consistent.
    \item[7.] \textbf{DeepGMM}\cite{bennett2019deep}: a variational method based on optimally-weighted GMM . We use its implementation in CausalML \footnote{https://github.com/CausalML/DeepGMM}. 
\end{itemize}

We compare our algorithm \textbf{AutoIV} with the following baseline methods: (1) \textbf{TrueIV:} use true valid IVs as a prior; (2) \textbf{RandIV:} use random variables (sampled from the same distribution of the true valid IVs) as IVs; (3) \textbf{UAS:}\cite{davies2015many} use equally weight to synthesize IVs from the IV candidates; (4) \textbf{WAS:}\cite{burgess2016combining} synthesize IVs by weighting the IV candidates based on their correlation to the treatment. 
We use the above methods to generate IVs (IV representations) and feed them to the downstream IV-based counterfactual prediction methods to testify the validity of the generated IVs (IV representations). 
To evaluate the performance of these IV synthesis methods under different IV candidates validity scenarios, we set: (1) \textbf{w/ $Z$:} parts of the valid IVs are given in the IV candidates, and (2) \textbf{w/o $Z$:} no valid IVs are given in the IV candidates. The latter setting is more practical in real-world applications and would make the task of synthesizing valid IVs (IV representations) more challenging as well as the IV-based counterfactual prediction.

\subsection{Low-dimensional Scenarios}
Similar to \cite{bennett2019deep}, we first implement experiments in low-dimensional scenarios (i.e., all the variables are in low-dimensional), and the data generating process is:
\begin{equation}
    \begin{aligned}
        &Y=g(X)+e+\sigma, \ X=Z_{1}+e+\gamma, \ Z\sim {\rm{Unif}}([-3,3]^{2}) \\
        &V=[Z;\gamma;\sigma], \qquad e\sim \mathcal{N}(0,1), \quad \ \ \ \gamma,\sigma\sim\mathcal{N}(0,0.1),
    \end{aligned}
\end{equation}
where $Z$ are the true valid IVs used as prior in the TrueIV baseline, while RandIV replaces it by randomly sampling from the same distribution of $Z$. $\sigma$ and $\gamma$ are noise. Variables $V$ are observed and used as the IV candidates which is composed by concatenating $Z$, $\gamma$, and $\sigma$. $e$ is an unobserved error term that is correlated to both the treatment $X$ and the outcome $Y$, $g$ is the true response function that chosen from the following settings (some are different from \cite{bennett2019deep} to increase the difficulty of counterfactual prediction):
\begin{equation}
    \begin{aligned}
        &\boldsymbol{\rm{step:}} \ g(X)=\begin{cases}-1 & X\ge0\\0&X<0\end{cases}\\
        &\boldsymbol{\rm{Linear:}} \ g(X)=-X \\
        &\boldsymbol{\rm{poly2d:}} \ g(X)=-0.1*X^2-0.4*X\\
        &\boldsymbol{\rm{poly3d:}} \ g(X)=0.05*X^3+0.1*X^2-0.8*X\\
        &\boldsymbol{\rm{abs:}} \ g(X)=|X| \\
    \end{aligned}
\end{equation}
We sample 500 samples for training, validation, and test, respectively. The values of $Z$, $X$, and $Y$ are standardized to avoid numerical problems. The representation dimensions of $Z$ and $C$ are set to the same, which is a hyper-parameter (the robustness of it is discussed in the later experiments). We plot the true and the estimated response function (i.e., $g$ and $\hat{g}$) in Figure \ref{fig-toy}. If the IVs fed in each method are more valid, the estimated response function would be more closer to the true response function (blue line). We find that (1) RandIV (orange line) fails badly in each case, while TrueIV achieves significantly better performance than RandIV, which indicates that IV information is necessary for removing confounding effect; (2) AutoIV (red line) achieves comparable or even better performance than TrueIV. It is may because AutoIV employs mutual information constraints as well as the representation calibration to further improve the IV representations validity, i.e., enhancing the relevance of the generated IV representations to the treatment and the exclusion to the outcome. 

To further improve the difficulty of the task, we then provide a more challenging data generating process by introducing confounders $C$:
\begin{equation}
    \begin{aligned}
        &Y=g(X)+C_{1...6}+e+\sigma, \ \  \ \ X=Z_{1...2}+C_{1...6}+e+\gamma\\
        &Z\sim {\rm{Unif}([-0.5,0.5]^2)}, \  \ \ \ \ \ \ C\sim {\rm{Unif}}([-0.5,0.5]^6)\\
        &e\sim \mathcal{N}(0,1), \ \ \  \gamma,\sigma\sim\mathcal{N}(0,0.1), \ \ \ V=[Z;C_{1:4};\gamma;\sigma] \ \
    \end{aligned}
    \label{DGP-equation}
\end{equation}
where $C_{1...6}$ denotes $C_1+C_2+...+C_6$, $Z_{1...2}$ denotes $Z_1+Z_2$. $C_{1:4}$ is used as a part of IV candidates for IV representation learning, while $C_{5:6}$ is directly employed for the downstream counterfactual prediction methods.
We report Mean Square Error (MSE) and standard error (Std) of the predicted counterfactual outcome over 20 runs in Table \ref{toy-table}. 
Similarly, we first find that RandIV performs poorly than TrueIV, indicating that valid IVs are important for removing confounding effect and accurate counterfactual prediction. 
Besides, the UAS, WAS, and AutoIV methods under w/ $Z$ setting achieve significantly better performance than w/o $Z$ setting, which is probably because the validity of the IV candidates allows IV synthesis methods to generate more valid IV representations. 
It is worth noting that most of the results under w/o Z setting with AutoIV method show better counterfactual prediction performance even compared with other methods under w/ Z setting. It suggests that AutoIV generates valid IV representations even there is no IV candidate is valid, and we attribute the success to the powerful ability of AutoIV in information control that makes the learned IV representations effectively satisfy the relevance and the exclusion conditions of the valid IVs for accurate counterfactual prediction.

Since representation dimension is a hyperparameter of the AutoIV algorithm, we design experiments by changing the representation dimension as 1, 2, 4, 8 (the true response function $g$ is set to $\rm{abs}$) and the results are shown in Figure \ref{fig-rep_dim}. 
We find that 2SLS (poly) and 2SLS (NN) is not robust enough to the changes of representation dimensions, which is may because their models are relatively simple.
We also see that DeepIV and KernelIV in both w/ $Z$ and w/o $Z$ settings are robust to the representation dimensions. While we note that DeepGMM method performs better in larger representation dimension setting, which is may because DeepGMM relies more heavily on parameter size, and higher dimensions bring more parameters in the fully-connected layer of the neural networks for DeepGMM.

AutoIV is a data-driven decomposed representation learning method, hence we implement experiments with different training data size settings ($g$ is set to $\rm{abs}$) as shown in Figure \ref{fig-num}.  It illustrates that AutoIV achieves great performance in different training data size settings. Moreover, larger data size will increase the decomposed representation learning performance and counterfactual prediction accuracy. 
However, it is not evident that the performance of other baseline methods is related to the training data size. 

We then give sensitivity analysis of the hyperparameters, i.e. $\alpha$ and $\eta$ in our algorithm. We show the performance of each method in the search space of each hyperparameter in Figure \ref{fig-hyperparameter}. It illustrates that in general the performance of our AutoIV algorithm is robust to $\alpha$ and $\eta$ with different downstream IV-based methods in counterfactual prediction.

To show the effectiveness of each part of the AutoIV algorithm, we conduct ablation studies by removing each component, including representation learning of $Z$ ($\mathcal{L}_{ZX}^{MI}+\mathcal{L}_{ZY}^{MI}$), representation learning of $C$ ($\mathcal{L}_{CX}^{MI}+\mathcal{L}_{CY}^{MI}$), decomposed regularization ($\mathcal{L}_{CZ}^{MI}$), and counterfactual prediction ($\mathcal{L}_{X}+\mathcal{L}_{Y}$). We implement the experiments ($g$ is set to $\rm{abs}$) on DeepIV, KernelIV, and DeepGMM, and the results are reported in Table \ref{table-ablation}. It shows that the necessity of each component in our AutoIV algorithm. Moreover, the two-stage procedure is shown important for further representation calibration. It is because mutual information constraints only control the information flow, but do not effectively enable them to be effective IV representations. While the general two-stage calibration process utilizes the gathered information to further synthesize powerful IV representations.

\subsection{High-dimensional Scenarios}
Following \cite{bennett2019deep}, we then implement experiments in high-dimensional scenarios with hand-written digit datasets MNIST \cite{lecun1998gradient}. To further testify the representation learning ability of AutoIV, we consider more complicated data composition that observed variables $V$ contain: (1) IVs $Z$, (2) confounders $F$ (i.e., variables that are related to $X$ and $Y$), (3) adjustments $A$ (i.e., variables that are only related to $Y$), (4) and unconcerned variables $U$ (i.e., variables that are independent of both the treatment $X$ and outcome $Y$). The data generating process is given as:
\begin{equation}\label{equ-DGP-high}
    \begin{aligned}
        &Y=g(X)+\mathbb{E}[F]+\mathbb{E}[A]+e, \quad X=\mathbb{E}[Z]+\mathbb{E}[F]+e\\
        &Z\sim {\mathcal{N}(0,1)^{dZ}}, \quad F\sim {\mathcal{N}(0,1)^{dF}}, \quad A\sim {\mathcal{N}(0,1)^{dA}} \\
        &U\sim {\mathcal{N}(0,1)^{dU}}, \quad C=[F,A,U], \quad V=[Z;C], \quad e\sim \mathcal{N}(0,1),
    \end{aligned}
\end{equation}
where $dZ$, $dF$, $dA$, $dU$ are the dimensions of $Z$, $F$, $A$, $U$ respectively.
Since UAS and WAS are only valid in the linear setting and are not competent to handle high-dimensional non-linear data, hence we compare RandIV, TrueIV, and AutoIV in the experiments of high-dimensional scenarios. 
Sine the non-linearity and high-dimension of data increase the difficulty of the task, we only consider w/ Z setting in these experiments. The response function $g$ is set to be $\rm{abs}$. We then give the following experimental settings:
\begin{small}    
 \begin{equation}
    \begin{aligned}
        &\boldsymbol{\rm{MNIST}_{Z}}: Z \stackrel{{\rm{Conv}}}{\longleftarrow} {\rm{MNSIT}_{rand}} \\
        &\boldsymbol{\rm{MNIST}_{C}}: C \stackrel{{\rm{Conv}}}{\longleftarrow} {\rm{MNSIT}_{rand}} \\
        &\boldsymbol{\rm{MNIST}_{ZC}}: Z \stackrel{{\rm{Conv}}}{\longleftarrow} {\rm{MNSIT}_{rand}}, C \stackrel{{\rm{Conv}}}{\longleftarrow} {\rm{MNSIT}_{rand}}
    \end{aligned}
    \label{MNIST-equation}
\end{equation}
\end{small}
where ${\rm{MNSIT}_{rand}}$ denotes randomly sampling from MNIST datasets. We adopt convolutional architecture (see Figure \ref{fig-conv}) to handle original MNIST images by following \cite{jason2017deepiv, bennett2019deep}, $Z$ and $C$ are sampled on the penultimate fully-connected (FC) layer with given dimensions.

We sample 1000 data points for training, validation, and test, respectively. We set 10, 10, 4, and 1 for $d_{Z}$, $d_{F}$, $d_{A}$, and $d_{U}$, respectively, and let representation dimension be 5 (AutoIV-5), 10 (AutoIV-10), 15 (AutoIV-15). The results are reported in Table \ref{table-mnist} with MSE and standard error of 20 runs. We find that the results of each method with AutoIV are significantly better than those with RandIV and superior to those with TrueIV. From the settings of AutoIV-5, AutoIV-10, and AutoIV-15, we see that the performance of AutoIV algorithm is robust to the change of representation dimension, showing its effectiveness in IV representation learning. 

We then analyze the performance of AutoIV with different dimensions of data composition and report the results in Table \ref{table-mnist-data}. 
It indicates that AutoIV is competent to generate valid IV representations in different data composition settings. All the experimental settings again show AutoIV's powerful representation learning ability in generating valid IV representation for accurate counterfactual prediction, which is even better than directly using the true valid IVs. 

Overall, these results highlight the great decomposed representation learning ability of our AutoIV algorithm in automatically generating the representation serving the role of IVs for accurate IV-based counterfactual prediction.

\section{Conclusions}\label{sec-con}
In this paper, we tackle the problem of decomposing and generating valid IV representations from the observed variables (i.e. the IV candidates). 
We relax the assumptions and conditions used by previous methods in handling this problem. 
We propose a novel Automatic Instrumental Variable decomposition (AutoIV) algorithm to decompose and learn valid representations of IVs automatically from the observed variables. 
We learn the IV representations by employing mutual information constraints, making the learned IV representations satisfy the conditions of the valid IVs in an adversarial game. 
Extensive empirical results in both low-dimensional and high-dimensional scenarios show the effectiveness of the AutoIV algorithm in generating IV representations and using them for IV-based counterfactual prediction with the downstream methods. 
The proposed AutoIV algorithm is an important addition to the toolkit of causal inference and IV-based counterfactual prediction.


\bibliographystyle{ACM-Reference-Format}
\bibliography{sample-base}










\end{document}